\documentclass[10pt,twocolumn,letterpaper]{article}

\usepackage[pagenumbers]{cvpr} 

\usepackage{booktabs}
\usepackage{graphicx}
\usepackage{pifont}
\usepackage{capt-of}

\newcommand{\cmark}{\textcolor{green}{\ding{51}}} 
\newcommand{\xmark}{\textcolor{red}{\ding{55}}} 

\usepackage{bbm}

\definecolor{cvprblue}{rgb}{0.21,0.49,0.74}
\usepackage[pagebackref,breaklinks,colorlinks,allcolors=cvprblue]{hyperref}

\def\paperID{} 
\def\confName{CVPR}
\def\confYear{2026}

\title{LTGS: Long-Term Gaussian Scene Chronology From Sparse View Updates}

\author{
Minkwan Kim$^{1}$ \quad
Seungmin Lee$^{1}$ \quad
Junho Kim$^{1}$ \quad
Young Min Kim$^{1,2}$ {\small \phantom{ }}\\
{\small $^{1}$ Dept. of Electrical and Computer Engineering, Seoul National University} \\
{\small $^{2}$ Interdisciplinary Program in Artificial Intelligence and INMC, Seoul National University}
}

\begin{document}
\twocolumn[{
\renewcommand\twocolumn[1][]{#1}
\maketitle
\begin{center}
    \centering
    \captionsetup{type=figure}
    \includegraphics[trim={0, 0, 0, 0}, clip, width=\linewidth]{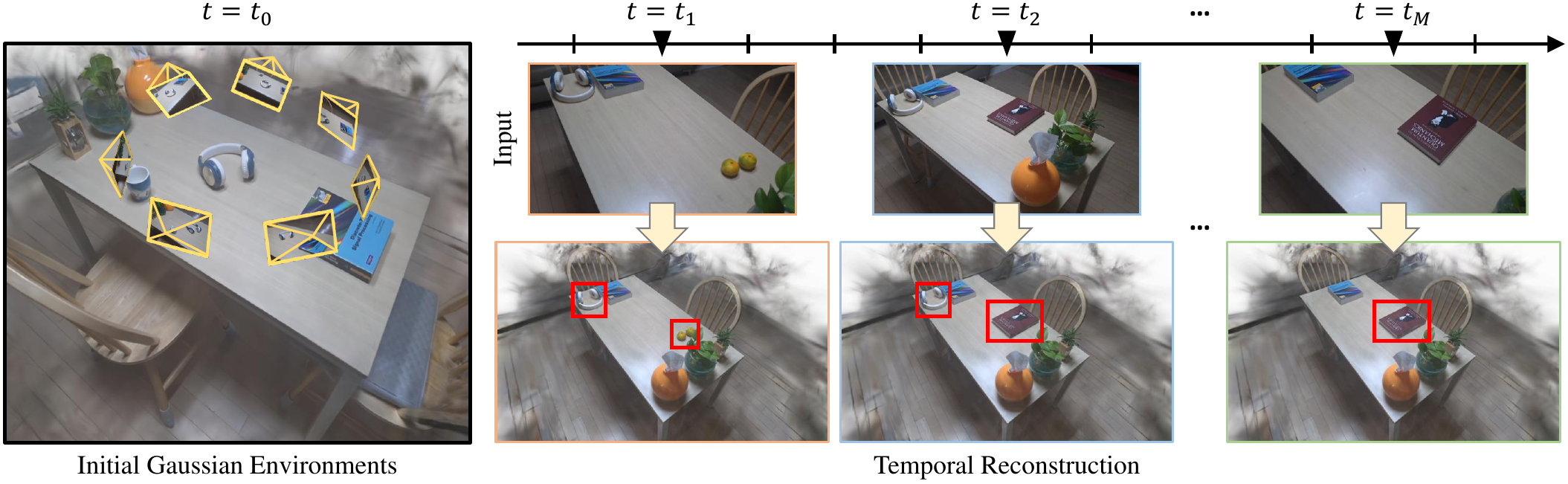}
    \captionof{figure}{\textbf{We introduce LTGS to efficiently update the Gaussian reconstruction of the initial environments.} Given the spatio-temporally sparse post-change images, our framework tracks object-level changes in 3D and models long-term scene evolution.}
    \label{fig:teaser}
\end{center}
}]
\maketitle
\begin{abstract}
Recent advances in novel-view synthesis can create the photo-realistic visualization of real-world environments from conventional camera captures.
However, the everyday environment experiences frequent scene changes, which require dense observations, both spatially and temporally, that an ordinary setup cannot cover.
We propose long-term Gaussian scene chronology from sparse-view updates, coined \textbf{LTGS}, an efficient scene representation that can embrace everyday changes from highly under-constrained casual captures. 
Given an incomplete and unstructured 3D Gaussian Splatting (3DGS) representation obtained from an initial set of input images, we robustly model the long-term chronology of the scene despite abrupt movements and subtle environmental variations. 
We construct objects as template Gaussians, which serve as structural, reusable priors for shared object tracks. Then, the object templates undergo a further refinement pipeline that modulates the priors to adapt to temporally varying environments given few-shot observations. 
Once trained, our framework is generalizable across multiple time steps through simple transformations, significantly enhancing the scalability for a temporal evolution of 3D environments.
As existing datasets do not explicitly represent the long-term real-world changes with a sparse capture setup, we collect real-world datasets to evaluate the practicality of our pipeline.
Experiments demonstrate that our framework achieves superior reconstruction quality compared to other baselines while enabling fast and light-weight updates. 
Project page is available at: \url{https://mkjjang3598.github.io/LTGS}.
\end{abstract}    
\vspace{-0.5em}

\section{Introduction}
\label{sec:intro}

With recent advances in novel-view synthesis, such as Neural Radiance Fields (NeRFs)~\cite{nerf} or 3D Gaussians Splatting (3DGS)~\cite{3dgs}, a casual user can reconstruct a 3D environment using a conventional camera input, and enjoy a photo-realistic experience of exploring the environment. 
The representation stores the complex distribution of light and geometry of the static scene in an unstructured format.
If we model the everyday environments where people live, daily activities often induce changes within the scene, quickly making the reconstruction obsolete.
One may need to rerun these algorithms from scratch, or incorporate 4D representations that encapsulate the dynamic movements of the 3D representation~\cite{d-nerf, 4dgs, dycheck}.
The former discards the previously acquired information, while the latter can only process with continuous observation of smooth motion.
Both approaches suffer from significant redundancy and are not desirable for modeling everyday environments in practical applications, such as location-based services, digital twins, or robotic setups.

We argue that a practical strategy for modeling evolving real-world environments is to efficiently detect and update changes.
Instead of requiring continuous captures of the entire scene, we suggest a light-weight update of the changed region from a sparse set of images, as demonstrated in~\cref{fig:teaser}.
While our setup suggests comparably flexible and realistic input requirements, it is highly under-constrained and therefore requires a strong scene prior.
The challenge is to efficiently update the scene without notable artifacts while maintaining the necessary information for photorealistic rendering.
Previous works that adapt novel-view synthesis into sparse input still suffer from severe artifacts when the viewpoint changes significantly~\cite{regnerf, sparsenerf, instantsplat}.
The continual learning method can sustain the pre-captured information~\cite{cl-nerf, cl-splats, gaussianupdate}, but relies on multiple captures for updates, as it lacks structural priors.

We propose an integrated pipeline that detects and updates Gaussian splatting scene representations in environments with diverse object states from sparse observations.
We refer to the framework as long-term Gaussian scene chronology from sparse view updates, or \textbf{LTGS}.
Real-world scenes can undergo multiple types of changes, including variations in geometry, appearance, or lighting.
Recent works demonstrate that subtle appearance and lighting changes can be partially addressed by adding learnable embeddings or auxiliary neural networks~\cite{nerf-w, neural_scene_chronology, wildgaussians, vastgaussian}.
In this work, we focus on \textbf{object-level changes}, which involve abrupt geometric alterations such as insertions, removals, replacements, or relocations, providing a structural mechanism to efficiently account for the consequences of daily interactions.

Given an initially reconstructed 3DGS without any segmentation, we need to robustly extract the object-level structure, which confines the granularity of change estimation under ambiguous observations.
Our scene update involves object tracking, relocalization, and reconstruction for individual objects.
By combining multiple image-space cues of segmentation and feature extraction, we detect and distill the change to a 3D representation.
We then aggregate the observations to build an object-level Gaussian template that models an object shared across time.
The template serves as a \textbf{reusable 3D prior} to relocalize objects at different times, resolving ambiguities in sparse views.
Then we reiterate the aggregation step such that the images of multiple time spans refine the shared template to best explain the overall observations via simple transformations.
To evaluate our framework in practical scenarios, we captured real-world datasets containing multiple shared objects in various layouts, with few-shot observations spanning multiple time steps. 
Our pipeline demonstrates robust performance in challenging scenarios where previous approaches struggle. Our key contributions are summarized as follows:
\begin{table}[t]
\centering
\resizebox{\linewidth}{!}{
    \begin{tabular}{l|ccccc}
        \toprule
        Method & Discont. motion & Temporal recon. & Few-shot & Speed \\
        \midrule
        3DGS~\citep{3dgs} & \xmark & \xmark & \xmark & Fast \\
        InstantSplat~\citep{instantsplat} & \xmark & \xmark & \cmark & Fast \\
        4DGS~\citep{4dgs} & \xmark & \cmark & \xmark & Moderate \\
        NSC~\citep{neural_scene_chronology} & \xmark & \cmark & \xmark & Slow \\
        3DGS-CD~\citep{3dgscd} & \cmark & \xmark & \cmark & Fast \\
        CL-NeRF~\citep{cl-nerf} & \cmark & \xmark & \xmark & Slow \\
        CL-Splats~\citep{cl-splats} & \cmark & \xmark & \xmark & Fast \\
        \midrule
        LTGS (Ours) & \cmark & \cmark & \cmark & Fast \\
        \bottomrule
    \end{tabular}
}
\caption{\textbf{Related methods comparison.} 
Our method captures abrupt geometric changes without requiring continuous motion and maintains reconstructions of multiple timesteps 
using a decomposable geometric scene prior that is reusable, thus allowing fast updates from a sparse set of images.
}
\label{table:related works}
\end{table}

\begin{itemize}
\item We address the problem of updating an initial 3DGS reconstruction in a highly efficient manner by using a set of spatio-temporally sparse images capturing long-term changes. 
\item We present LTGS, an integrated strategy to track, associate, and relocalize the objects, and reconstruct the evolving scenes.
\item  We propose a new real-world dataset, casually capturing environments with dynamic object-level changes across multiple timesteps to evaluate our framework.  
\end{itemize}

\section{Related Works}
\label{sec:related_work}

We build on Gaussian splatting representation and propose a novel yet challenging practical setup that allows lightweight updates of temporal changes from sparse view inputs. The setting partially shares important properties with recent variations of novel-view synthesis that enable temporal extensions or few-shot reconstruction, as summarized in~\cref{table:related works}.

\subsection{Non-static scene reconstruction}
Several works successfully extend NeRFs~\cite{d-nerf, t-nerf} or 3DGS~\cite{dynamicgaussians,4dgs,free-timeGS} to model dynamic scenes. These works 
require dense input observations both temporally and spatially~\cite{dycheck}, which can be challenging to obtain in practice. Furthermore, the reconstructed dynamics recover slow, continuous motions~\cite{d-nerf, t-nerf, dynamicgaussians, 4dgs, free-timeGS} or color variations on quasi-static geometry~\cite{neural_scene_chronology}.
Another line of research integrates a continual learning framework, transforming the initial reconstruction to match gradual changes over time.
The extension of 3DGS~\cite{cl-splats, gaussianupdate} is inherently faster in rendering compared to NeRFs~\cite{cl-nerf, clnerf}, and therefore enjoys faster training times.
However, these works still require more than ten input images to adapt to scene changes and eventually lose information from previous time steps.

\subsection{Few-shot NeRF and Gaussian splatting}
Several works pioneered relieving the dense-view requirement of NeRFs or 3DGS by incorporating geometric priors or regularization techniques~\cite{regnerf, sparsenerf, FSGS, scade}.
Recent geometric vision foundation models such as MASt3R~\cite{mast3r} provide strong geometric priors, enabling effective scene representation when combined with Gaussian splatting~\cite{instantsplat, splatt3r}, particularly in cases where conventional structure-from-motion (SfM)~\cite{colmap_sfm} fails.
However, in the context of updating scene changes, these methods cannot preserve the initial reconstruction, which leads to significant performance degradation, such as severe floating artifacts in novel views.
Our approach instead builds a geometric structure of reusable priors by aggregating actual observations, and sustains consistent long-term temporal reconstruction despite a sparse set of temporal image captures.

\subsection{Change detection and segmentations in 3DGS}
While change detection has remained a long-standing problem in computer vision~\cite{cscdnet, the_change_you_want_to_see}, vision foundation models facilitate more generalizable ways to detect changes.
Recent approaches~\cite{gescf, hasdanythingchanged, 3dgscd} estimate change regions by leveraging the encoded feature embeddings of the segment anything model (SAM)~\cite{sam}.
Similarly, DINO features~\cite{dinov2} are leveraged to detect change regions~\cite{cl-splats, change_gaussians}.
Recently, several approaches have demonstrated that one can separate 3D objects from 3DGS with 2D masks~\cite{flashsplat, cob-gs}.
Our framework incorporates a similar framework to build Gaussian templates of the shared objects in evolving 3D environments.

\section{Method}
\label{sec:method}

\subsection{Overview}
\label{subsec:overview}
Given the initial 3DGS reconstruction of the scene $\mathcal{G}_0$, our goal is to update the scene from a set of images $\mathcal{I}=\{ I^i\}_t$, where $t \in \{ t_1, t_2, \ldots, t_M\}$ represents a sparsely sampled time stamps, and the input images for each time stamp are captured from a small number of viewpoints $i=1, \ldots, N_t$ that are not fixed beforehand. 
Our goal is to acquire the temporal evolution of the scene $\mathcal{S} = \{\mathcal{G}_0, \mathcal{G}_1, ... \mathcal{G}_{M}\}$ that recovers its states at sampled time steps in the input.
The representations are a set of Gaussian splats, and each splat element is parameterized as $\{\mu, q, s, \alpha, c\}$, where $\mu \in \mathbb{R}^{N\times3}$ denotes 3D center position, $q \in \mathbb{R}^{N\times4}$ represents quaternion representation of orientation (we denote its conversion into a rotation matrix as $R$), $s \in \mathbb{R}^{N\times3}$ specifies scale, $\alpha \in \mathbb{R}^{N\times1}$ denotes opacity, and $c \in \mathbb{R}^{N\times48}$ encodes view-dependent color using third-order spherical harmonics.

\begin{figure*}[t]
    \centering
    \includegraphics[width=\linewidth]{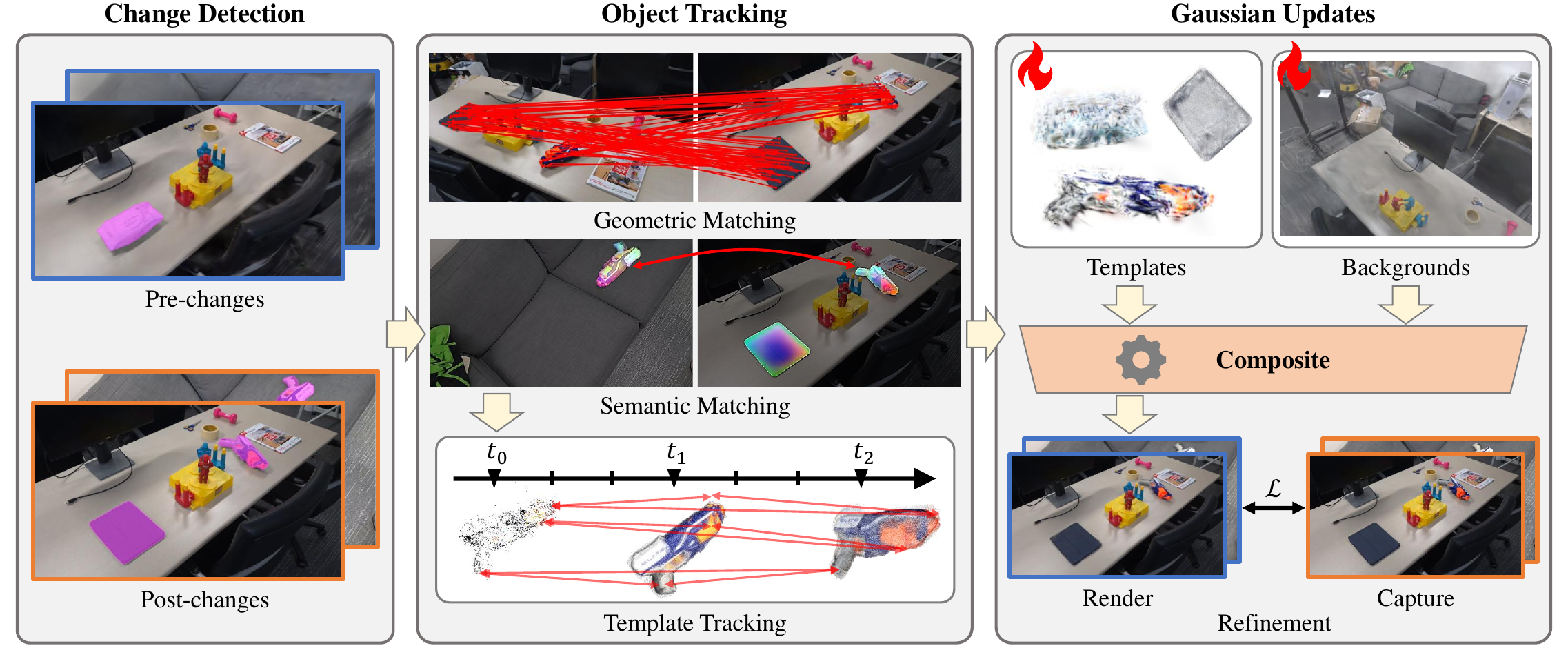}
    \caption{\textbf{Method overview.} We propose an integrated pipeline to update an initial reconstruction given the collection of post-change captures. Our pipeline first estimates the camera poses of the input capture and compares them against renderings of the initial reconstruction in the same view to detect object-level changes. 
    We aggregate detected objects from multiple viewpoints and timestamps to create 3D Gaussian templates, and finally update the temporal scenes by compositing the templates at their respective states with the background.}
    \vspace{-1.0em}
    \label{fig:method_overview}
\end{figure*}

Assuming the scene is an everyday environment with frequent interactions, it may experience geometric changes due to object displacements.
We focus on modeling object-level movements and develop a decomposed representation that can serve as a strong geometric prior despite sparse observations, as shown in~\cref{fig:method_overview}.
We develop a light-weight pipeline that i) performs foundation model inference~\cite{mast3r, sam} for dense local matches and instance identification, and ii) \textit{reuses} these results for subsequent steps in instance association, coarse geometry estimation, and scene updating.
Specifically, we robustly detect object-level changes (\cref{subsec:change_detection}) and collect information for individual objects from the input image set (\cref{subsec:object_tracking}). We separately train Gaussian templates for the objects and static background (\cref{subsec:GS_opt}), such that blending them with the correct transform can best match the input images at the corresponding time step.
While we incorporate strong cues of foundation models, they do not directly translate into stable 3D updates. 
Our aggregation strategy is critical for converting their per-view predictions into coherent object-level templates.
Overall, our framework can achieve high-quality reconstruction in temporally evolving scenes while suppressing prominent artifacts with minimal input and computation.

\subsection{Change detection} 
\label{subsec:change_detection}
%
We first find the exact camera positions of $\{I^i\}_t$ despite local changes, such that we can compare the observations from different time stamps against the initial Gaussian reconstruction $\mathcal{G}_0$.
We use a robust hierarchical localization pipeline~\cite{hloc} and render the initial reconstruction in the same viewpoints $\{\hat{I}^i\}_t$.

To decompose our scene representation into movable objects and static background, we detect temporal changes.
We incorporate both semantic and photometric criteria to identify the differences between the rendered $\{\hat{I}^i\}_t$ and the captured $\{I^i\}_t$ images.
Semantic differences detect object-level changes despite lighting variations and other adversaries, and are measured by the cosine similarity of SAM features~\cite{sam}, similar to recent change detection methods~\cite{gescf, hasdanythingchanged, 3dgscd}.
Photometric differences are evaluated using the structural similarity index measure (SSIM), which can detect subtle object deviations not observable by SAM.
The pixel-wise differences of the combined criteria are binarized by a scene-specific threshold chosen by the statistics~\cite{otsu_threshold}, resulting in an initial pseudo mask for changed regions.

Note that we can robustly extract object-level changes from sparse input with the semantic masks.
Specifically, we select the object region to be the set of SAM masks that sufficiently overlap with the pseudo masks while containing semantically dissimilar features compared to the initial image, as the result of change~\cite{gescf}.
The aid of SAM masks effectively ignores differences due to floating artifacts in the rendering images and reliably extracts object regions.
The resulting detected masks are dilated by 3 pixels 
to maintain sufficient information for 3D aggregation across multi-view observations, mitigating the impact of pixel-wise errors or slight misalignments. See the supplementary material for detailed implementation details on change detection.
The overall pipeline produces stable outputs, even in the presence of potential inaccuracies as discussed in~\cref{sec:experiments}.

\subsection{Object tracking and template reconstruction}
\label{subsec:object_tracking}
After obtaining change masks
, we match the changed objects within the individual images and extract initial 3D Gaussian templates for them.
We then refine the relative transforms between the 3D templates and the image observations, which can serve as input to integrated optimization for 3D scene $\mathcal{S}$.
The entire process takes only 30 seconds to track and reconstruct templates for five discrete timesteps.

\vspace{-1.0em}
\paragraph{2D instance matching}
While one can associate object masks using low-level image features such as SIFT~\cite{sift}, our sparse setting makes it challenging to match small objects with few discriminative features.
Also, objects are sometimes dynamic in appearance or geometry, which further complicates the process.
To address these issues, we combine the strength of both dense geometric features from MASt3R~\cite{mast3r} and semantic features obtained from SAM~\cite{sam} in~\cref{subsec:change_detection}. 
We first match multi-view images incorporating MASt3R features~\cite{mast3r} for all pairs of images within the same time stamp $\{I^i \}_t$.
Using the output, we build a graph where the nodes are object masks and the edges are matches with pairwise correspondences. 
By examining the graph structure, we can assign instance IDs to the matched components and filter out unmatched objects, such that we can naturally overcome artifacts.
The intra-timestep matching then provides reliable starting points for establishing matchings across different timesteps.
We aggregate the SAM features within the object region, which have been computed to detect changes in~\cref{subsec:change_detection},  and build a matrix that records extensive pairwise cosine similarity of the aggregated features.
Based on the matrix, we leverage Hungarian matching~\cite{algorithms} to match object instances.

\vspace{-1.0em}
\paragraph{Gaussian object template extraction}
After associating 2D change masks, we are ready to build object-level Gaussian templates, which can be refined and function as a reusable geometric prior.
We first decompose the initial Gaussian reconstruction $\mathcal{G}_0$ into a set of template objects $\mathcal{T}_0 = \{o_{0,k}\}$ and the background $\mathcal{B}_0$.
We formulate it as a segmentation problem for individual splats, and solve for the optimal label assignment as proposed in~\cite{flashsplat}.
Objects in a later sequence do not exist in $\mathcal{G}_0$, and we initialize the templates $\{\mathcal{T}_t|t > 0\}$ using 3D point clouds estimated with MASt3R.
%
Note that we reuse the extracted descriptors of MASt3R in previous stages and optimize per-view depth maps with fixed camera parameters from~\cref{subsec:change_detection} to reconstruct 3D point clouds.
%
The point clouds directly provide the positions and colors of the Gaussian splats, and we simply initialize uniform opacity, identity rotation, and uniform scaling as~\cite{instantsplat}.
For the background, we maintain a single global set of Gaussians $\mathcal{B}_0$. 
When occluded regions in the initial reconstruction become visible after changes, we augment this background representation by initializing the newly observed areas using point maps from MASt3R.

\vspace{-1.0em}
\paragraph{Gaussian object template tracking}
After initializing 3D templates, we deduce temporal states of the objects by tracking their movements and verifying consistency in 3D.
For the object instances matched from different time steps, we compare their 3D overlap using robust point cloud registration.
However, the MASt3R points or Gaussian reconstructions are incomplete and noisy with irregular density, and often cannot be registered using conventional point cloud pipelines such as ICP~\cite{icp} or RANSAC-based approaches~\cite{ransac}.
Instead, we establish correspondences by augmenting DINO features~\cite{dinov2} to each point and apply a robust point cloud registration pipeline~\cite{teaserpp} to register templates.
The registration yields 6DoF poses between pairs of template points $P_{t\rightarrow{\tilde{t}},k}=\{R_{t\rightarrow{\tilde{t}},k}, T_{t\rightarrow{\tilde{t}},k}\}$ such that we can assess the geometric consistency across the temporal track by thresholding with the Chamfer distance~\cite{chamfer}.
If the points are close enough, we avoid redundancy by selecting a single 3D template per matched instance, along with its relative transforms over time.
Since we conservatively select 
templates, we can naturally represent an object under significant geometric variations as different instances without modifying the framework, as shown in~\cref{fig:articulation}.
We further refine the shared template in the next stage. 
See the supplementary material for details on object tracking and object-level template reconstruction.

\subsection{Long-term Gaussian splats optimization}
\label{subsec:GS_opt}
While the selected templates can explain the other time steps with sufficient geometric overlap, they are derived from a single time step and can be noisy and incomplete, especially in different time steps with significant viewpoint changes. 
We aggregate the collection of observations and optimize the parameters of Gaussian splats (\cref{subsec:overview}).
The templates from initial time step can be transformed into a time step $t$ using the registration parameters  $P_{0\rightarrow{t},k}=\{R_{0\rightarrow{t},k},T_{0\rightarrow{t},k}\}$ from~\cref{subsec:object_tracking} as following:
\vspace{-0.5em}
\begin{equation}
\begin{split}
(\mu_{t,k}, R_{t,k}, c_{t,k}) = (
    \mu_{0,k} R_{0\rightarrow{t},k}^\top + T_{0\rightarrow{t},k}^\top, \\
R_{0\rightarrow{t},k} R_{0,k},\; c_{0,k}\, \mathcal{R}_{\text{SH}}(R_{0\rightarrow{t},k})^\top
).
\end{split}
\end{equation}
Here, $R_{0,k}$ and $R_{t,k}$ are rotation matrices corresponding to $q_{0,k}$ and $q_{t,k}$ respectively and SH coefficients are rotated via rotation operator $\mathcal{R}_{\text{SH}}$~\cite{Gaussreg}.
In addition, we apply a temporal opacity filter per object $\mathcal{M}_{t,o}$ such that transient objects become invisible (zero opacity).
Since the 6DoF poses obtained from~\cref{subsec:object_tracking} are not precise at the pixel level, we additionally set the 6DoF poses of the object templates as an optimization parameter.

These strategies enable efficient modeling of long-term environments, but they also risk overfitting the initial assets to post-change views.
To address this issue, we additionally leverage training camera poses used at the initial stages to render the scene and enforce consistency in the rendered images. 
Formally, we can render and update the image from the $i$th viewpoint at time $t$ by defining the optimization problem as follows:
\vspace{-0.5em}
\begin{equation}
\label{eq:optimization}
\begin{split}
\hat{I}_t^i = \operatorname{Rasterize}(\mu_{t,k}, q_{t,k}, s_{t,k}, \mathcal{M}_{t,o}\cdot \alpha_{t,k}, c_{t,k}), \\
\{\mu^*, q^*, s^*, \alpha^*, c^*, P^*\} = \arg\min \, \mathcal{L}_{\text{photo}}\!\left(\hat{I}_t^i,{I}_t^i \right),
\end{split}
\end{equation}
where $I_t^i$ contains both captured post-change images and renderings from the initial camera poses.
For $\mathcal{L}$, we use the standard L1 loss with D-SSIM loss that was used in the original implementation of 3DGS~\cite{3dgs}.
The background scene is initialized with $\mathcal{B}_0$ and similarly refined using all the information from different times. 
As the initial templates provide a reasonable approximation, 5000 iterations are sufficient to refine the parameters without densifying or cloning Gaussians, and we also skip the opacity resetting stages.
Once optimized, our framework easily scales to multiple timesteps by simple transformations of template Gaussians.

\section{Experiments}
\label{sec:experiments}
\subsection{Datasets \& baselines}
\paragraph{Datasets} 
We use a synthetic dataset from CL-NeRF~\cite{cl-nerf}, which contains three scenes captured at different timesteps: [\textsc{Whiteroom, Kitchen, Rome}]. 
Each timestep includes object-level sequential operations, such as addition, deletion, replacement, and movement.
However, the motions are simple and do not exhibit diverse variations between objects in different steps.
%
We additionally captured challenging real-world scenes, where objects may abruptly reappear in different configurations.
%
The dataset consists of image collections captured in five scenes at 5 different timesteps: [\textsc{Cafe, Diningroom, Hall, Lab, Livingroom}].

\vspace{-1.0em}
\paragraph{Baselines} 
We compare our work to recent NeRFs and 3DGS variants.
First, we evaluate against the original (1) 3DGS~\cite{3dgs} using all images for different timesteps as a reference.
We further compare with (2) InstantSplat~\cite{instantsplat}, a few-shot reconstruction method applied independently at each timestep. To account for frameworks explicitly modeling dynamic scenes, we include (3) 4DGS~\cite{4dgs} and (4) Neural Scene Chronology (NSC)~\cite{neural_scene_chronology}.
In addition, we compare (5) 3DGS-CD, which explicitly detects and updates object-level changes.
Finally, we evaluate against continual learning frameworks, including (6) CL-NeRF~\cite{cl-nerf} and (7) CL-Splats~\cite{cl-splats}.
The works and their capabilities are also summarized in~\cref{table:related works}.
Refer to the supplement for details.

\begin{figure*}[t]
    \centering
    \includegraphics[width=1\linewidth]{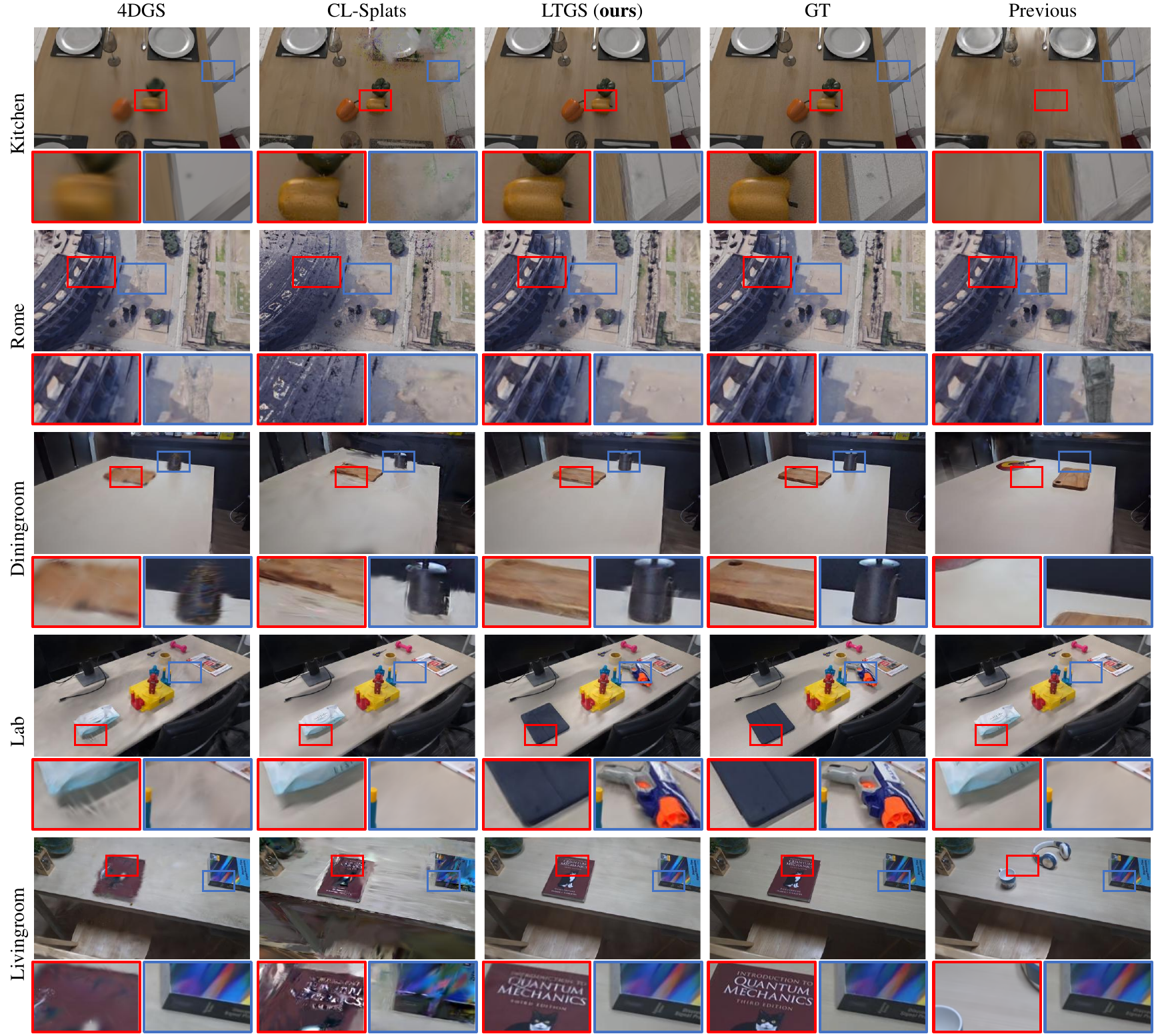}
    \caption{\textbf{Qualitative comparisons of our method.} We illustrate the results of our method using the CL-NeRF dataset and our dataset.
}\label{fig:qualitative_comparisons}
\vspace{-1.0em}
\end{figure*}

\subsection{Comparative studies}
We evaluate our method and baselines on novel-view synthesis tasks across scenes with multiple timesteps.
For every setting, we use three images at each timestep from various angles to capture scene changes.
Our framework largely outperforms baselines both qualitatively and quantitatively, as demonstrated in~\cref{table:quantitative_comparison} and~\cref{fig:qualitative_comparisons}. 
It successfully reconstructs diverse object-level changes while remaining robust to limitations imposed by sparse views.
In particular, InstantSplat~\cite{instantsplat} is designed for fast and lightweight reconstruction specifically in a few-shot setting, and thus cannot maintain its performance on a free-viewpoint setting covering the full scene.
4DGS~\cite{4dgs} and NSC~\cite{neural_scene_chronology} struggle to precisely model the discrete changes, such as added or removed objects.
The snapshots in~\cref{fig:qualitative_comparisons} also include overly smooth results in the change regions.

While 3DGS-CD~\cite{3dgscd} also quickly handles object-level placement changes, our approach consistently outperforms, as our approach better accounts for added and removed objects.
Recent continual-learning frameworks also struggle to address spatio-temporally sparse settings. 
CL-NeRF~\cite{cl-nerf} performs well on synthetic datasets but cannot track complex real-world changes. 
Also, the implicit representation of CL-NeRF occasionally results in degraded sharpness, as shown in the appendix.
The optimization of CL-Splats~\cite{cl-splats} fails to maintain its stability in the sparse-view inputs, as its mask estimation becomes unreliable.
The effects can also be observed in the \textsc{Lab} scene (4th row) in~\cref{fig:qualitative_comparisons}.


\begin{table*}[t]
\centering
\resizebox{0.8\linewidth}{!}{
    \begin{tabular}{l|cccc|cccc}
	\toprule
	{} & \multicolumn{4}{c|}{CL-NeRF dataset (synthetic)} & \multicolumn{4}{c}{Our dataset (real)} \\ 
        {Method} & PSNR $\uparrow$ & SSIM $\uparrow$ & LPIPS $\downarrow$ & Time $\downarrow$ & PSNR $\uparrow$ & SSIM $\uparrow$ & LPIPS $\downarrow$ & Time $\downarrow$ \\
        \midrule
        3DGS~\citep{3dgs} & 24.53 & 0.789 & 0.392 & 6 min & 19.56 & 0.857 & 0.272 & 8 min
\\
        InstantSplat~\citep{instantsplat} & 18.98 & 0.601 & 0.466 & 3 min & 19.36 & 0.785 & 0.343 & 3 min \\
        4DGS~\citep{4dgs} & 26.13 & 0.786 & 0.411 & 24 min & 21.49 & 0.850 & 0.322 & 29 min \\
        NSC~\citep{neural_scene_chronology} & 20.63 & 0.698 & 0.465 & $>$10 hours & 17.52 & 0.755 & 0.439 & $>$10 hours \\
        3DGS-CD~\citep{3dgscd} & 23.61 & 0.727 & 0.437 & \textbf{2 min} & 20.94 & 0.774 & 0.348 & \textbf{2 min} \\
        CL-NeRF~\citep{cl-nerf} & 25.53 & 0.730 & 0.465 & 2 hours & 20.95 & 0.815
 & 0.379 & 2 hours\\
        CL-Splats~\citep{cl-splats} & 25.84 & 0.772 & 0.416 & 3 min & 21.12 & 0.829 & 0.312 & 3 min \\
        \midrule
        LTGS (ours) & \textbf{27.17} & \textbf{0.795} & \textbf{0.376} & 6 min & \textbf{23.46} & \textbf{0.889} & \textbf{0.230} & 7 min \\
        \bottomrule
    \end{tabular}
}
\caption{\textbf{Quantitative comparisons on CL-NeRF dataset and our dataset.} We compared our method against NeRF-based and Gaussian splatting variants. The best results are highlighted in \textbf{bold}.}
\vspace{-1.0em}
\label{table:quantitative_comparison}
\end{table*}

We also report the total time spent reconstructing different timesteps in~\cref{table:quantitative_comparison}.
In our framework, the total processing time is approximately 6.5 minutes (2.5 min for change detection, 0.5 min for instance matching, 3.5 min for 5000 iteration updates), measured on an NVIDIA RTX 4090.
In conclusion, while prior methods either oversmooth dynamic changes or fail under sparse inputs, our framework achieves accurate and efficient reconstruction of evolving scenes with strong object-level consistency.

\begin{table}[tb]
\small
\centering
\resizebox{0.85\linewidth}{!}{
    \begin{tabular}{l|ccc}
        \toprule
        Configuration & PSNR $\uparrow$ & SSIM $\uparrow$ & LPIPS $\downarrow$ \\
        \midrule
        w/o Obj. Tracking & 23.26 & 0.885 & 0.234 \\ 
        w/o Pose Opt. & 23.33 & 0.886 & 0.232 \\
        w/o BG Init. & 23.29 & 0.885 & 0.233 \\ 
        w/o Training View & 23.11 & 0.885 & 0.240 \\ 
        \midrule
        Full (ours) & \textbf{23.46} & \textbf{0.889} & \textbf{0.230} \\ 
        \bottomrule
    \end{tabular}
}
\caption{\textbf{Ablation study}. We demonstrate the effect of different optimization configurations.} 
\vspace{-0.5em}
\label{table:ablation}
\end{table}

\begin{figure}
    \centering
    \includegraphics[width=\linewidth]{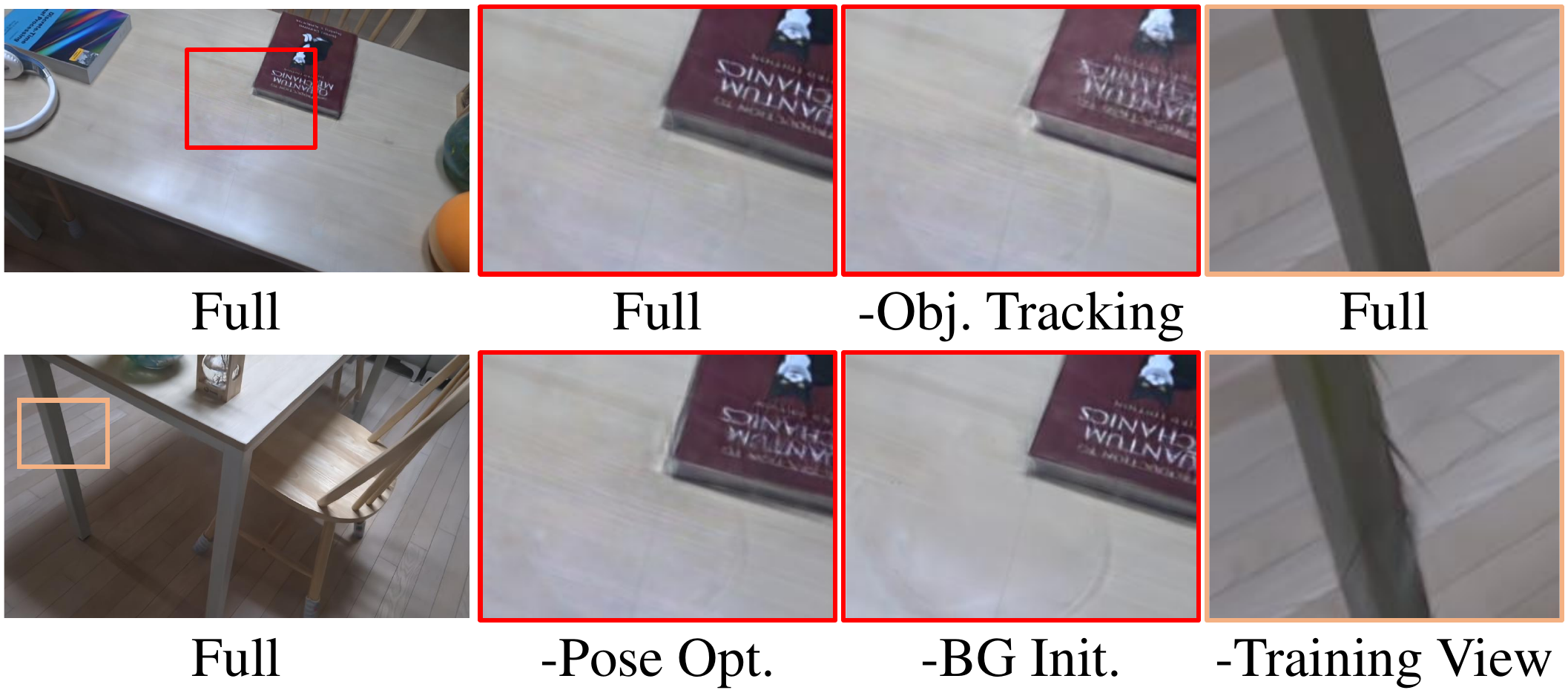}
    \caption{\textbf{Visual comparison of ablation study.} We visualize the effect of each components listed in~\cref{table:ablation}.}
    \vspace{-1.0em}
    \label{fig:ablation_study}
\end{figure}

\subsection{Ablation studies}
We conducted ablation studies on various components of the method, as demonstrated in~\cref{table:ablation}.
Since our components are primarily designed to improve the modeling of changing objects rather than the overall quality of the initially reconstructed scenes, the image quality metric itself did not fully reveal significant improvement. 
Accordingly, we additionally demonstrate the qualitative comparisons focusing on the change regions ~\cref{fig:ablation_study}.

We first tested the effect of instance matching, where we removed the template association step and built new object-level Gaussians at every time step.
Without using object Gaussian templates as a reusable prior, we could not handle sparse view limitations, leaving traces of removed objects as shown in~\cref{fig:ablation_study}.
6DoF pose updates also increased the reconstruction quality, as they account for pixel-level errors from subtle pose errors after registration.

While our primary ablations focus on reconstructing dynamic objects, we also examined the impact of background initialization. 
Specifically, when objects disappear and leave previously occluded regions visible, initializing these empty areas with MASt3R~\cite{mast3r} point clouds alleviates background artifacts. 
As shown in~\cref{fig:ablation_study}, the joint use of global background Gaussians with this initialization strategy further reduces artifacts after object removal.
Including training views of initial timesteps also enhanced the quality, as reported in~\cref{table:ablation}. 
As we only leverage a few-shot images, some unseen regions, such as under the table or back of the chair in~\cref{fig:ablation_study}, include sharp artifacts, which degrade the rendering of several viewpoints.
We verified that each component clearly affected the enhancement of both object and background reconstructions without compromising the initial reconstruction.

\begin{figure}[t]
    \centering
    \includegraphics[width=\linewidth]{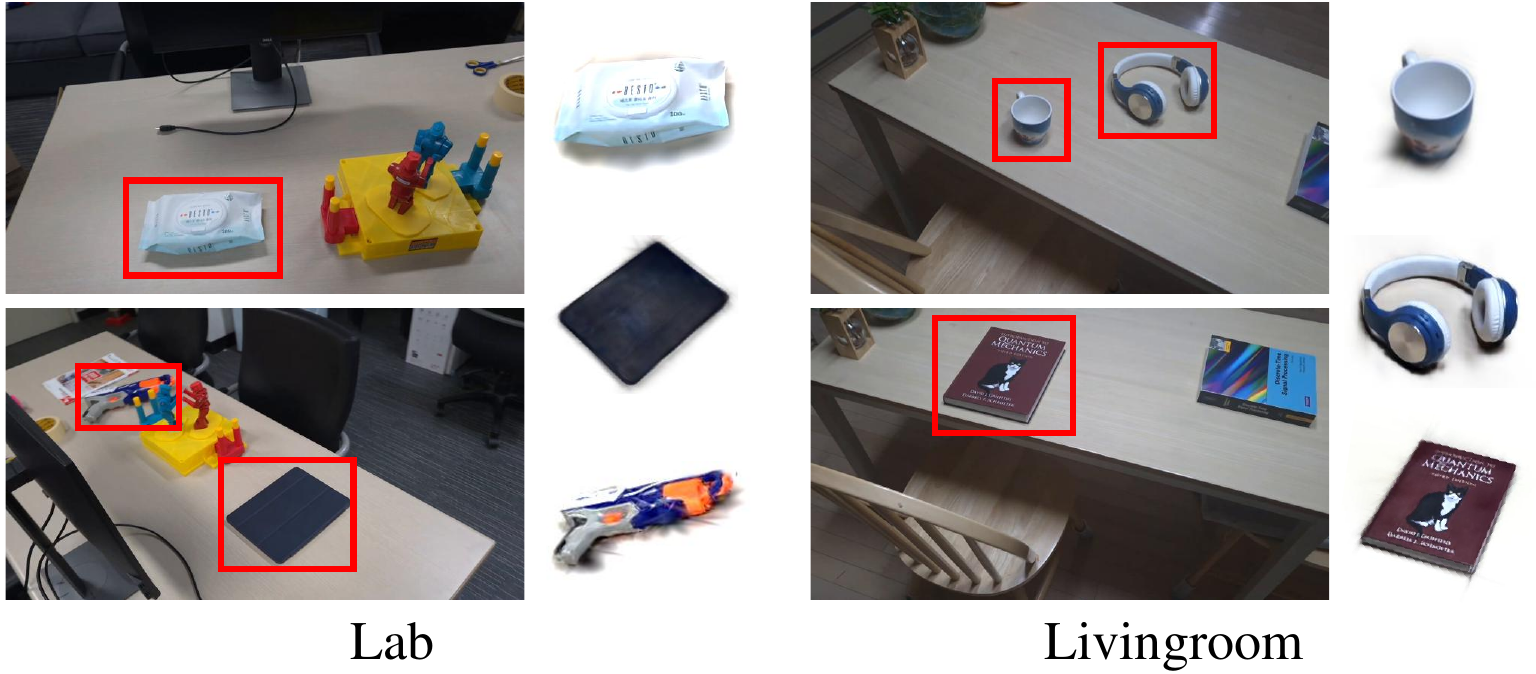}
    \caption{\textbf{Object template visualization.} We sampled several captures from the initial state and post-change captures and corresponding object-level Gaussian templates.}
    \vspace{-1.0em}
    \label{fig:object_templates}
\end{figure}

\subsection{Performance Analysis}
\paragraph{Object-level reconstruction.} 
After optimization, our framework produces object-level reconstruction, where learned Gaussian templates can be directly leveraged for scene composition and temporal reasoning.
To illustrate this, we visualize the optimized object-level Gaussian templates in~\cref{fig:object_templates}. 
We selected some viewpoints from initial and post-change captures, and rasterized the object-level Gaussians to the corresponding viewpoints.
Our framework effectively disentangles individual objects from the scene while preserving consistent geometry and appearance across time steps.
Notably, even with few-shot observations, the optimized object templates exhibit well-defined shapes without severe artifacts along object boundaries. 
This suggests that the optimization effectively integrates multi-timestep cues into coherent object-level representations.
Such clean object templates also highlight the potential of our method for modeling object-level changes in scenes with longer temporal variances.

\begin{figure}[t]
    \centering
    \vspace{-0.5em}
    \includegraphics[width=1\linewidth]{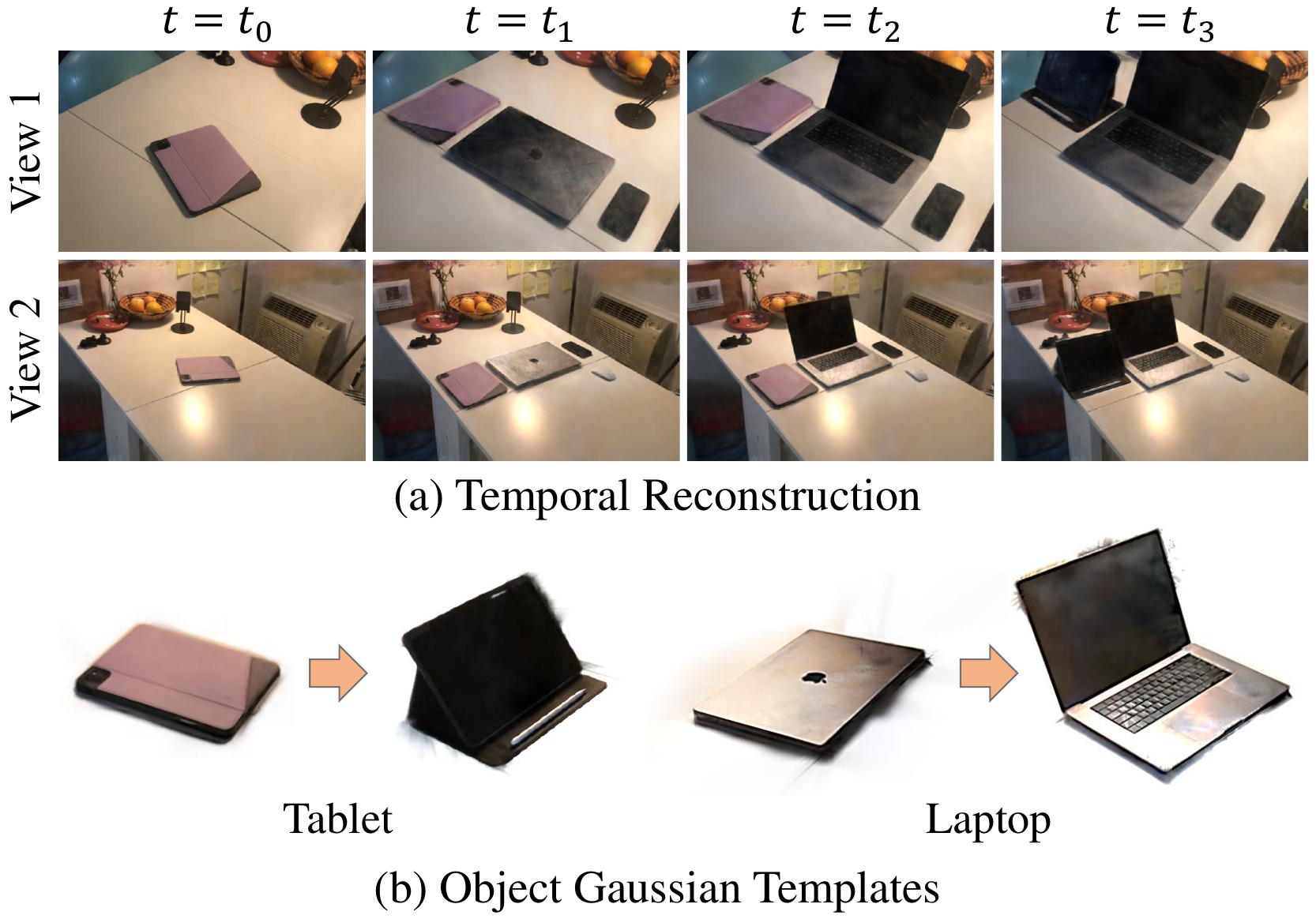}
    \caption{\textbf{Challenging real-world scenarios.} We demonstrate (a) reconstruction results with articulations and (b) a visualization of object-level reconstructions.}
    \vspace{-1.0em}
    \label{fig:articulation}
\end{figure}

\paragraph{Non-rigid transformations or articulations.}
\vspace{-1.0em}
We further tested our framework on more challenging setups as shown in~\cref{fig:articulation}. 
In real-world scenarios, object-level changes often involve non-rigid transformations or articulations.
Our method handles such cases by defining separate object Gaussian templates for objects in different articulation states.
Although these templates are not rigidly tracked, they are all identified as a single object instance due to our robust 2D matching pipeline.
We visualize the reconstructed scene with temporal variations and object-level reconstructions across different states for the \textsc{MAC} scene from the world-across-time (WAT) dataset introduced in CLNeRF~\cite{clnerf}. 
Thus, our method provides a principled way to represent objects in different states through independent templates.
Moreover, the tracked object templates can be combined with recent works~\cite{screwsplat}, modeling object articulation for more detailed tracking, which is left as future work.

\begin{figure}[t]
    \centering
    \vspace{-0.5em}
    \includegraphics[width=1\linewidth]{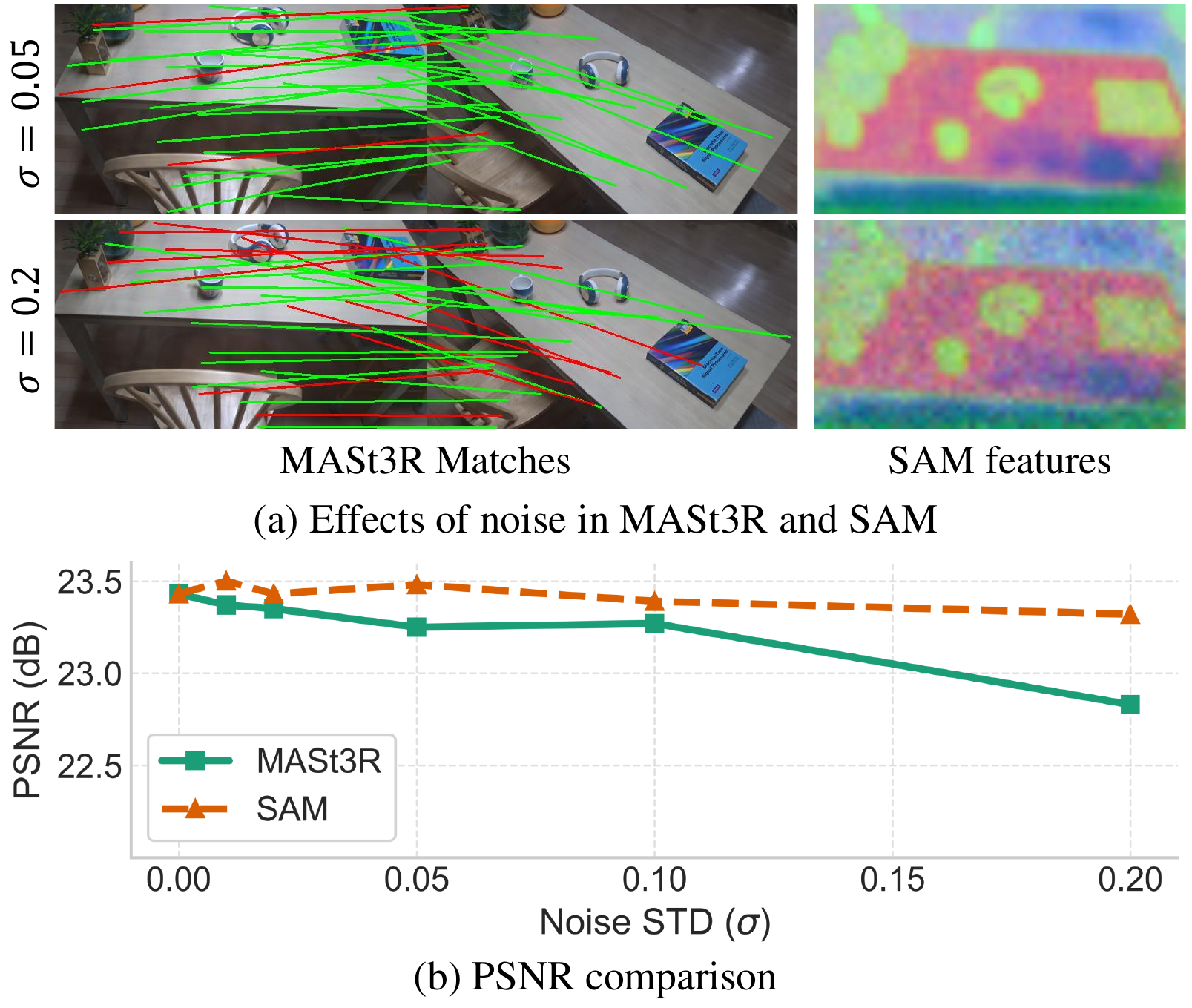}
    \caption{\textbf{Robustness evaluation.} We demonstrate (a) the effects of varying noise levels on components of our pipeline, and (b) the reconstruction results measured in PSNR.}
    \vspace{-1.0em}
    \label{fig:robustness}
\end{figure}

\paragraph{Robustness to errors of foundation models}
\vspace{-1.0em}
We conducted experiments to verify the robustness of our framework against potential errors occurred from foundation models. 
Specifically, we injected random noise with varying standard deviations into the MASt3R~\cite{mast3r} descriptors and SAM~\cite{sam} embeddings as demonstrated in~\cref{fig:robustness}.
Even with the perturbed MASt3R descriptors and SAM embeddings, our framework achieved reliable reconstruction without severe degradation, with stable instance matching and template reconstruction.
The contour-based aggregation helps smooth out local perturbations, and the graph-based instance matching across multiple timesteps further enhances robustness to errors from pretrained models.
Moreover, the errors in reconstructed Gaussian templates caused by noisy MASt3R descriptor can be compensated by the proposed long-term optimization process in~\cref{subsec:GS_opt}.

\section{Conclusion} 
\vspace{-0.5em}
\label{sec:conclusion}
We present LTGS, an integrated framework for modeling scenes with long-term changes given spatiotemporally sparse images.
Our strategy stably builds and exploits object-centric templates under the challenging setup.
Several comparative studies and ablation studies have verified that the combination of our components significantly outperforms the baselines.
We further verified our frameworks to be applicable to several extensions, such as object-level reconstruction and reconstructing more challenging setups with non-rigid transformations or articulations.
In conclusion, we believe this framework offers a promising foundation for building a coherent structural representation that is reusable for a long temporal horizon.
As our framework primarily targets scenes with geometric variations, it poses challenges for scenes with significant lighting changes or severe appearance changes with fixed geometries, such as monitors, if these are not captured as changes. 
We leave the problem of modeling the heavy lighting changes with shadows for future work.

\paragraph{Acknowledgements}
\vspace{-0.5em}
This work was supported by IITP grant funded by the Korea government (MSIT) (RS-2023-00216821, Development of Beyond X-verse Core Technology for Hyper-realistic interactions by Synchronizing the Real World and Virtual Space) and the BK21 FOUR program of the Education and Research Program for Future ICT Pioneers, Seoul National University in 2026.
\newpage
{
    \small
    \bibliographystyle{ieeenat_fullname}
    \bibliography{main}

@String(CVPR= {IEEE Conf. Comput. Vis. Pattern Recog.})

@String(ICCV= {Int. Conf. Comput. Vis.})

@String(ECCV= {Eur. Conf. Comput. Vis.})

@String(TOG= {ACM Trans. Graph.})

@String(CVPR  = {CVPR})

@String(ICCV  = {ICCV})

@String(ECCV  = {ECCV})

@String(TOG   = {ACM TOG})

@article{llff,
  title={Local Light Field Fusion: Practical View Synthesis with Prescriptive Sampling Guidelines},
  author={Ben Mildenhall and Pratul P. Srinivasan and Rodrigo Ortiz-Cayon and Nima Khademi Kalantari and Ravi Ramamoorthi and Ren Ng and Abhishek Kar},
  journal={ACM Transactions on Graphics (TOG)},
  year={2019},
}

@inproceedings{nerf,
  title={NeRF: Representing Scenes as Neural Radiance Fields for View Synthesis},
  author={Ben Mildenhall and Pratul P. Srinivasan and Matthew Tancik and Jonathan T. Barron and Ravi Ramamoorthi and Ren Ng},
  year={2020},
  booktitle={ECCV},
}

@inproceedings{scade,
      title = {SCADE: NeRFs from Space Carving with Ambiguity-Aware Depth Estimates},
      author = {Mikaela Angelina Uy and Ricardo Martin-Brualla and Leonidas Guibas and Ke Li},
      booktitle = {Conference on Computer Vision and Pattern Recognition (CVPR)},
      year = {2023}
}

@InProceedings{regnerf,
          author    = {Michael Niemeyer and Jonathan T. Barron and Ben Mildenhall and Mehdi S. M. Sajjadi and Andreas Geiger and Noha Radwan},  
          title     = {RegNeRF: Regularizing Neural Radiance Fields for View Synthesis from Sparse Inputs},
          booktitle = {Proc. IEEE Conf. on Computer Vision and Pattern Recognition (CVPR)},
          year      = {2022},
}

@article{sparsenerf,
    title={SparseNeRF: Distilling Depth Ranking for Few-shot Novel View Synthesis},
    author={Guangcong and Zhaoxi Chen and Chen Change Loy and Ziwei Liu},
    journal={IEEE/CVF International Conference on Computer Vision (ICCV)},
    year={2023}
}

@article{cl-nerf,
  title={CL-NeRF: continual learning of neural radiance fields for evolving scene representation},
  author={Wu, Xiuzhe and Dai, Peng and Deng, Weipeng and Chen, Handi and Wu, Yang and Cao, Yan-Pei and Shan, Ying and Qi, Xiaojuan},
  journal={Advances in Neural Information Processing Systems},
  volume={36},
  pages={34426--34438},
  year={2023}
}

@inproceedings{clnerf,
title={CLNeRF: Continual Learning Meets NeRF},
author={Zhipeng Cai, Matthias Müller},
year={2023},
booktitle={ICCV},
}

@inproceedings{d-nerf,
    title={{D-NeRF: Neural Radiance Fields for Dynamic Scenes}},
    author={Pumarola, Albert and Corona, Enric and Pons-Moll, Gerard and Moreno-Noguer, Francesc},
    booktitle={Proceedings of the IEEE/CVF Conference on Computer Vision and Pattern Recognition},
    year={2020}
}

@inproceedings{t-nerf,
  title={Neural 3d video synthesis from multi-view video},
  author={Li, Tianye and Slavcheva, Mira and Zollhoefer, Michael and Green, Simon and Lassner, Christoph and Kim, Changil and Schmidt, Tanner and Lovegrove, Steven and Goesele, Michael and Newcombe, Richard and others},
  booktitle={Proceedings of the IEEE/CVF conference on computer vision and pattern recognition},
  pages={5521--5531}, 
  year={2022}
}

@inproceedings{dycheck,
    title={Monocular Dynamic View Synthesis: A Reality Check},
    author={Gao, Hang and Li, Ruilong and Tulsiani, Shubham and Russell, Bryan and Kanazawa, Angjoo},
    booktitle={NeurIPS},
    year={2022},
}

@inproceedings{neural_scene_chronology,
  title={Neural Scene Chronology},
  author={Lin, Haotong and Wang, Qianqian and Cai, Ruojin and Peng, Sida and Averbuch-Elor, Hadar and Zhou, Xiaowei and Snavely, Noah},
  booktitle={CVPR},
  year={2023}
}

@inproceedings{nerf-w,
author = {Martin-Brualla, Ricardo
        and Radwan, Noha
        and Sajjadi, Mehdi S. M.
        and Barron, Jonathan T.
        and Dosovitskiy, Alexey
        and Duckworth, Daniel},
title = {{NeRF in the Wild: Neural Radiance Fields for
       Unconstrained Photo Collections}},
booktitle = {CVPR},
year={2021}
}

@Article{3dgs,
      author       = {Kerbl, Bernhard and Kopanas, Georgios and Leimk{\"u}hler, Thomas and Drettakis, George},
      title        = {3D Gaussian Splatting for Real-Time Radiance Field Rendering},
      journal      = {ACM Transactions on Graphics},
      number       = {4},
      volume       = {42},
      month        = {July},
      year         = {2023},
      url          = {https://repo-sam.inria.fr/fungraph/3d-gaussian-splatting/}
}

@inproceedings{FSGS,
  title={Fsgs: Real-time few-shot view synthesis using gaussian splatting},
  author={Zhu, Zehao and Fan, Zhiwen and Jiang, Yifan and Wang, Zhangyang},
  booktitle={European conference on computer vision},
  pages={145--163},
  year={2024},
  organization={Springer}
}

@article{wildgaussians,
  title={{W}ild{G}aussians: {3D} Gaussian Splatting in the Wild},
  author={Kulhanek, Jonas and Peng, Songyou and Kukelova, Zuzana and Pollefeys, Marc and Sattler, Torsten},
  journal={NeurIPS},
  year={2024}
}

@inproceedings{vastgaussian,
  title     = {VastGaussian: Vast 3D Gaussians for Large Scene Reconstruction},
  author    = {Lin, Jiaqi and Li, Zhihao and Tang, Xiao and Liu, Jianzhuang and Liu, Shiyong and Liu, Jiayue and Lu, Yangdi and Wu, Xiaofei and Xu, Songcen and Yan, Youliang and Yang, Wenming},
  booktitle = {CVPR},
  year      = {2024}
}

@article{flashsplat,
  title={FlashSplat: 2D to 3D Gaussian Splatting Segmentation Solved Optimally},
  author={Shen, Qiuhong and Yang, Xingyi and Wang, Xinchao},
  journal={European Conference of Computer Vision},
  year={2024}
}

@inproceedings{cob-gs,
  title     = {COB-GS: Clear Object Boundaries in 3DGS Segmentation Based on Boundary-Adaptive Gaussian Splitting},
  author    = {Zhang, Jiaxin and Jiang, Junjun and Chen, Youyu and Jiang, Kui and Liu, Xianming},
  booktitle = {CVPR},
  year      = {2025}
}

@inproceedings{Gaussreg,
  title={Gaussreg: Fast 3d registration with gaussian splatting},
  author={Chang, Jiahao and Xu, Yinglin and Li, Yihao and Chen, Yuantao and Feng, Wensen and Han, Xiaoguang},
  booktitle={European Conference on Computer Vision},
  pages={407--423},
  year={2024},
  organization={Springer}
}

@inproceedings{dynamicgaussians,
  title={Dynamic 3D Gaussians: Tracking by Persistent Dynamic View Synthesis},
  author={Luiten, Jonathon and Kopanas, Georgios and Leibe, Bastian and Ramanan, Deva},
  booktitle={3DV},
  year={2024}
}

@InProceedings{4dgs,
    author    = {Wu, Guanjun and Yi, Taoran and Fang, Jiemin and Xie, Lingxi and Zhang, Xiaopeng and Wei, Wei and Liu, Wenyu and Tian, Qi and Wang, Xinggang},
    title     = {4D Gaussian Splatting for Real-Time Dynamic Scene Rendering},
    booktitle = {Proceedings of the IEEE/CVF Conference on Computer Vision and Pattern Recognition (CVPR)},
    month     = {June},
    year      = {2024},
    pages     = {20310-20320}
}

@InProceedings{free-timeGS,
    author    = {Wang, Yifan and Yang, Peishan and Xu, Zhen and Sun, Jiaming and Zhang, Zhanhua and Chen, Yong and Bao, Hujun and Peng, Sida and Zhou, Xiaowei},
    title     = {FreeTimeGS: Free Gaussian Primitives at Anytime Anywhere for Dynamic Scene Reconstruction},
    booktitle = {Proceedings of the IEEE/CVF Conference on Computer Vision and Pattern Recognition (CVPR)},
    month     = {June},
    year      = {2025},
    pages     = {21750-21760}
}

@misc{instantsplat,
        title={InstantSplat: Sparse-view Gaussian Splatting in Seconds},
        author={Zhiwen Fan and Kairun Wen and Wenyan Cong and Kevin Wang and Jian Zhang and Xinghao Ding and Danfei Xu and Boris Ivanovic and Marco Pavone and Georgios Pavlakos and Zhangyang Wang and Yue Wang},
        year={2024},
        eprint={2403.20309},
        archivePrefix={arXiv},
        primaryClass={cs.CV}
}

@article{splatt3r,
  title={Splatt3r: Zero-shot gaussian splatting from uncalibrated image pairs},
  author={Smart, Brandon and Zheng, Chuanxia and Laina, Iro and Prisacariu, Victor Adrian},
  journal={arXiv preprint arXiv:2408.13912},
  year={2024}
}

@inproceedings{cscdnet,
  title={Weakly supervised silhouette-based semantic scene change detection},
  author={Sakurada, Ken and Shibuya, Mikiya and Wang, Weimin},
  booktitle={2020 IEEE International conference on robotics and automation (ICRA)},
  pages={6861--6867},
  year={2020},
  organization={IEEE}
}

@misc{the_change_you_want_to_see,
      title={The Change You Want to See}, 
      author={Ragav Sachdeva and Andrew Zisserman},
      year={2022},
      eprint={2209.14341},
      archivePrefix={arXiv},
      primaryClass={cs.CV},
      url={https://arxiv.org/abs/2209.14341}, 
}

@article{3dgscd,
  title={3DGS-CD: 3D Gaussian Splatting-Based Change Detection for Physical Object Rearrangement},
  author={Lu, Ziqi and Ye, Jianbo and Leonard, John},
  journal={IEEE Robotics and Automation Letters},
  year={2025},
  publisher={IEEE}
}

@inproceedings{cl-splats,
  author    = {Ackermann, Jan and Kulhanek, Jonas and Cai, Shengqu and Haofei, Xu and Pollefeys, Marc and Wetzstein, Gordon and Guibas, Leonidas and Peng, Songyou},
  title     = {CL-Splats: Continual Learning of Gaussian Splatting with Local Optimization},
  booktitle = {Proceedings of the IEEE/CVF International Conference on Computer Vision (ICCV)},
  year      = {2025}
}

@article{gaussianupdate,
  title={GaussianUpdate: Continual 3D Gaussian Splatting Update for Changing Environments},
  author={Zeng, Lin and Zhao, Boming and Hu, Jiarui and Shen, Xujie and Dang, Ziqiang and Bao, Hujun and Cui, Zhaopeng},
  journal={arXiv preprint arXiv:2508.08867},
  year={2025}
}

@InProceedings{change_gaussians,
    author    = {Galappaththige, Chamuditha Jayanga and Lai, Jason and Windrim, Lloyd and Dansereau, Donald and Sunderhauf, Niko and Miller, Dimity},
    title     = {Multi-View Pose-Agnostic Change Localization with Zero Labels},
    booktitle = {Proceedings of the Computer Vision and Pattern Recognition Conference (CVPR)},
    month     = {June},
    year      = {2025},
    pages     = {11600-11610}
}

@article{screwsplat,
  title={ScrewSplat: An End-to-End Method for Articulated Object Recognition},
  author={Kim, Seungyeon and Ha, Junsu and Kim, Young Hun and Lee, Yonghyeon and Park, Frank C},
  journal={arXiv preprint arXiv:2508.02146},
  year={2025}
}

@inproceedings{colmap_sfm,
    author={Sch\"{o}nberger, Johannes Lutz and Frahm, Jan-Michael},
    title={Structure-from-Motion Revisited},
    booktitle={Conference on Computer Vision and Pattern Recognition (CVPR)},
    year={2016},
}

@inproceedings{hloc,
  title     = {From Coarse to Fine: Robust Hierarchical Localization at Large Scale},
  author    = {Paul-Edouard Sarlin and
               Cesar Cadena and
               Roland Siegwart and
               Marcin Dymczyk},
  booktitle = {CVPR},
  year      = {2019}
}

@article{sift,
  title={Distinctive image features from scale-invariant keypoints},
  author={Lowe, David G},
  journal={International journal of computer vision},
  volume={60},
  number={2},
  pages={91--110},
  year={2004},
  publisher={Springer}
}

@misc{mast3r,
      title={Grounding Image Matching in 3D with MASt3R}, 
      author={Vincent Leroy and Yohann Cabon and Jerome Revaud},
      booktitle = {ECCV},
      year = {2024}
}

@InProceedings{gescf,
    author    = {Kim, Jae-Woo and Kim, Ue-Hwan},
    title     = {Towards Generalizable Scene Change Detection},
    booktitle = {Proceedings of the Computer Vision and Pattern Recognition Conference (CVPR)},
    month     = {June},
    year      = {2025},
    pages     = {24463-24473}
}

@article{hasdanythingchanged,
  title={Has anything changed? 3d change detection by 2d segmentation masks},
  author={Adam, Aikaterini and Karantzalos, Konstantinos and Grammatikopoulos, Lazaros and Sattler, Torsten},
  journal={arXiv preprint arXiv:2312.01148},
  year={2023}
}

@article{otsu_threshold,
  title={A threshold selection method from gray-level histograms},
  author={Otsu, Nobuyuki and others},
  journal={Automatica},
  volume={11},
  number={285-296},
  pages={23--27},
  year={1975}
}

@article{sam,
  title={Segment Anything},
  author={Kirillov, Alexander and Mintun, Eric and Ravi, Nikhila and Mao, Hanzi and Rolland, Chloe and Gustafson, Laura and Xiao, Tete and Whitehead, Spencer and Berg, Alexander C. and Lo, Wan-Yen and Doll{\'a}r, Piotr and Girshick, Ross},
  journal={arXiv:2304.02643},
  year={2023}
}

@misc{dinov2,
  title={DINOv2: Learning Robust Visual Features without Supervision},
  author={Oquab, Maxime and Darcet, Timothée and Moutakanni, Theo and Vo, Huy V. and Szafraniec, Marc and Khalidov, Vasil and Fernandez, Pierre and Haziza, Daniel and Massa, Francisco and El-Nouby, Alaaeldin and Howes, Russell and Huang, Po-Yao and Xu, Hu and Sharma, Vasu and Li, Shang-Wen and Galuba, Wojciech and Rabbat, Mike and Assran, Mido and Ballas, Nicolas and Synnaeve, Gabriel and Misra, Ishan and Jegou, Herve and Mairal, Julien and Labatut, Patrick and Joulin, Armand and Bojanowski, Piotr},
  journal={arXiv:2304.07193},
  year={2023}
}

@article{teaserpp,
  title={{TEASER: Fast and Certifiable Point Cloud Registration}},
  author={H. Yang and J. Shi and L. Carlone},
  journal={{IEEE} Trans. Robotics},
  Year = {2020} 
}

@article{dfs,
  title={Depth-first search and linear graph algorithms},
  author={Tarjan, Robert},
  journal={SIAM journal on computing},
  volume={1},
  number={2},
  pages={146--160},
  year={1972},
  publisher={SIAM}
}

@article{algorithms,
  title={Algorithms for the assignment and transportation problems},
  author={Munkres, James},
  journal={Journal of the society for industrial and applied mathematics},
  volume={5},
  number={1},
  pages={32--38},
  year={1957},
  publisher={SIAM}
}

@InProceedings{chamfer,
author = {Fan, Haoqiang and Su, Hao and Guibas, Leonidas J.},
title = {A Point Set Generation Network for 3D Object Reconstruction From a Single Image},
booktitle = {Proceedings of the IEEE Conference on Computer Vision and Pattern Recognition (CVPR)},
month = {July},
year = {2017}
}

@inproceedings{icp,
  title={Method for registration of 3-D shapes},
  author={Besl, Paul J and McKay, Neil D},
  booktitle={Sensor fusion IV: control paradigms and data structures},
  volume={1611},
  pages={586--606},
  year={1992},
  organization={Spie}
}

@article{ransac,
  title={Random sample consensus: a paradigm for model fitting with applications to image analysis and automated cartography},
  author={Fischler, Martin A and Bolles, Robert C},
  journal={Communications of the ACM},
  volume={24},
  number={6},
  pages={381--395},
  year={1981},
  publisher={ACM New York, NY, USA}
}
}
\


\maketitlesupplementary
\appendix

\section{Implementation details}
\subsection{Change detection} 
\label{suppl_sec:change_detection}
We detect fine object-level changes in 2D by using the semantic prior of the SAM model~\cite{sam}.
We prepare a pair of images ($I_t^i,\hat{I}_t^i$) from captured and rendered images from the corresponding viewpoints. 
To find abstract differences between the rendered and captured images, we extract features from the segmentation model for both images.
We used features obtained from the pretrained SAM encoder and interpolated them to match the original image resolution, after which we computed pairwise cosine similarities for comparison.
We additionally use SSIM to capture structural differences, where using only semantic cosine similarity struggles to find slightly deviated objects.

To obtain coarse change masks, we need a binarization by thresholding the obtained differences. 
Since the binary coarse mask is highly sensitive to manually defined thresholds, which vary significantly across scenes, we adopt Otsu’s method~\cite{otsu_threshold} to automatically determine the threshold $\tau_{cos}$.
We obtain the coarse binary masks $\mathcal{M}_{t, \text{coarse}}^i$ as follows: 
\begin{equation}
\begin{aligned}
    M_{t,\text{coarse}}^i &= \gamma \cdot \text{cos}(\mathcal{E}(I_t^i),\mathcal{E}(\hat{I}_t^i)) \\
    &+ (1-\gamma) \cdot \text{SSIM}(I_t^i, \hat{I}_t^i) \le \tau_{\text{cos}},
\end{aligned}
\label{suppl_eq:coarse_mask}
\end{equation}
where $\mathcal{E}$ denotes the feature extractor of SAM. $\gamma=0.7$ was used throughout our experiments.

After obtaining coarse change regions, we extract fine-grained object-level change masks to model instance-wise changes.
Since precise object-level changes are difficult to capture using the coarse stages described above, we leverage the automatic mask generation from SAM~\cite{sam} within the coarse binary mask $M_{t, \text{coarse}}^i$.
For each generated mask, we first calculate the intersection-over-union (IoU) between its region and the coarse binary mask. 
We further compare the cosine similarity between the accumulated features within those regions in the extracted features of the rendered and captured images.
The fine object-level masks can be obtained as follows:
\begin{equation}
\begin{aligned}    
\mathcal{O}^i_t = & \bigcup_{k} \Big\{ o_{t,k}^i \Big| 
\mathrm{IoU}(o_{t,k}^i, M^{i}_{t,\text{coarse}}) \ge \tau_{\mathrm{IoU}} \land \\
& \cos\!\big(\Phi(o_{t,k}^i,\mathcal{E}(I_t^i)), \Phi(o_{t,k}^i, \mathcal{E}(\hat I_t^i))\big) \le \tau_{\mathrm{cos}} \Big\},
\end{aligned}
\label{suppl_eq:fine_mask}
\end{equation}
where $o_{t,k}^i$ denotes the $k$th object mask generated by automatic mask generation, and $\Phi(m, X) = \frac{1}{|m|} \sum_{p \in \Omega} m(p)\,X(p)$ denotes the average pooling operator of feature $X$ within the interior $\Omega$ of mask $m$.
We select and keep only the highly overlapped and semantically different masks following~\cite{gescf}.
We further removed masks occupying only small regions to prevent noise, and we dilated the change masks to address pixel-wise errors.

\subsection{Object tracking}
\label{suppl_sec:object_tracking}
We provide additional details of our pipeline to associate with changed instances. 
Given 2D object-level change masks from Sec. 3.2 of the main paper, we leverage both visual feature and SAM feature matching to associate 2D change masks.
As mentioned in Sec. 3.3 of the main paper, we separate the instance matching into intra-timestep matching and cross-timestep matching.
Let $\mathcal{O}_t^i = \{ o_{t,1}^i, o_{t,2}^i, \dots, o_{t,N_o}^i \}$ be the set of object-level masks of $i$th viewpoint at timestep $t$. 
We first compute pairwise matches across images $I_t^i$ and $I_t^j$ using MASt3R within the same timestep. 
Given the extracted MASt3R descriptors $\mathbf{d}_{t,k}^i$ and $\mathbf{d}_{t,l}^j$  for object $o_{t,k}^i$ and $o_{t,l}^j$ respectively, we find descriptor matches $\mathcal{M}_t^{(i,k)\leftrightarrow(j,l)}$ within object change masks as follows:
\begin{equation}
\begin{aligned}    
\mathcal{M}_t^{(i,k)\leftrightarrow(j,l)} &= \text{match}(\mathbf{d}_{t,k}^i, \mathbf{d}_{t,l}^j), \\
o_{t,k}^i \in \mathcal{O}_t^i, &\; o_{t,l}^j \in \mathcal{O}_t^j, \; i \neq j.
\end{aligned}
\end{equation}
Note for matching, we follow the fast reciprocal matching procedure from MASt3R~\cite{mast3r}.
Based on these matches, we construct a graph $G_t = (N_t, E_t)$ as:

\begin{equation}
\begin{split}
N_t &= \bigcup_i \mathcal{O}_t^i, \\
E_t = \big\{
    \big(o_{t,k}^i,\, o_{t,l}^j,&\, 
    |\mathcal{M}_t^{(i,k)\leftrightarrow(j,l)}|\big)
    \;\big|\;
    i \neq j
\big\}.
\end{split}
\end{equation}

where the nodes $N_t$ are objects and the edge weight $E_t$ encodes the total number of matches between every pair of objects $o_{t,k}^i$ and $o_{t,l}^j$.  
We then cluster the graph using the depth-first search (DFS) algorithm~\cite{dfs} to identify connected components, 
where each component corresponds to a unique global object identity. 
Here, we retain an edge only if the number of matches exceeds a threshold $\tau_{match}$. 
Based on these clustering results, we assign instance IDs to the matched components and filter out unmatched objects for consistency.
This filtering strategy gives robustness to detected instances that are inaccurate due to the artifacts in rendered images.

After obtaining object-level matches for every sequence within an identical timestep, it is essential to associate object masks across different timesteps.
For each object $o_{t,k}^i$ and $o_{\tilde t,l}^j$ where $t$ and $\tilde t$ are the set of target times after intra-timestep matching, we accumulate the SAM~\cite{sam} features in the object region, and build a matrix $\mathbf{S}_{k\leftrightarrow l}$ that contains cosine similarity among every pair as follows:
\begin{equation}
\mathbf{S}_{k\leftrightarrow l} = \cos\!\big(\Phi(o_{t,k}^i,\mathcal{E}(I_t^i)), \Phi(o_{\tilde t,l}^j,\mathcal{E}(I_{\tilde t}^j))\big). 
\end{equation}
Based on the matrix, we leverage Hungarian matching~\cite{algorithms} to solve an optimal assignment problem between the instances as $\pi^* = \arg\max_{\pi} \sum_{k} \mathbf{S}_{k\leftrightarrow  \pi(k)}$,
where $\pi(k)$ denotes the matched object in timestep $\tilde t$.
After semantic matching, we filter out false pairs for those with the cosine similarities lower than $\tau_{cos}$ defined in~\cref{suppl_eq:coarse_mask} and~\cref{suppl_eq:fine_mask}.
Note that we conduct this process identically for every possible timestep pair for  $t\in[0,T]$. 

\subsection{Object Gaussian Template reconstruction}
\label{suppl_sec:template_reconstruction}
Given the tracked object masks $M=\{M_t^i|i=1,...,N_v; t=0,...,T\}$, we provide additional details of the construction and initialization of object-level Gaussian Splats.
Here, we define the total number of instances at the initial timestep as $E$. 
For the objects that emerge in initial reconstruction, we separate those using the optimal label assignment problem introduced in FlashSplat~\cite{flashsplat}.
For our task, the problem is defined as follows:
\begin{equation}
\label{suppl_eq:optimal_label_assignment}
\begin{aligned}
\min_{\{P_k\}} 
\mathcal{F} = \sum_{i} & \left| 
\sum_{k} P_k\alpha_{k}T_{k} - M_0^i
\right|, \\
P_k & \in \{0,1,...,E\},
\end{aligned}
\end{equation}
where $\alpha_{k}, T_{k}$ each denotes the alpha value and transmittance during volume rendering, and $P_k$ denotes the per-Gaussian 3D label. 
Among the $P_k$, index 0 corresponds to the background, while the remaining indices correspond to the foreground.
The above equation solves the problem of assigning the 3D label $P_k$ by volume rendering them to the image domain to match the given multiview masks at $t=0$.
Specifically, we use the majority voting algorithm as follows:
\begin{equation}
\label{suppl_eq:majority_voting}
\begin{aligned}
P_{k} &= \arg\max_{n \in \{0,m\}} A_n, \\
A_m &= \sum_{i}\ \alpha_{k}T_{k}\mathbbm{1}(M_0^i, m), \\
A_0 &= \sum_{i} \sum_{e \neq m} \alpha_{k}T_{k}\mathbbm{1}(M_0^i, e),
\end{aligned}
\end{equation}
where $\mathbbm{1}(M_0^i,m)$ denotes the indicator function which is equal to 1 if the pixel in mask $M_0^i$ belongs to object $m$, and 0 otherwise.
~\cref{suppl_eq:majority_voting} solves the assignment problem by allocating the label that maximizes the weighted contribution of Gaussians within the object mask regions.
Please refer to the original FlashSplat~\cite{flashsplat} paper regarding the details and the derivation.
We additionally filter Gaussians that are out of the object mask region after directly projecting the centers to remove floating artifacts. 

After registration and geometric verification as presented in Sec. 3.3 of the main paper, we initialize Gaussians for objects that do not exist in the initial reconstruction.
We first extract point clouds for new objects from the global scene reconstruction of MASt3R~\cite{mast3r} with estimated poses from the hierarchical localization pipeline~\cite{hloc}.
In the original implementation of MASt3R, camera parameters were optimized jointly with per-view depth maps and global scales.  
We modify the optimization loop to operate only on depth maps with scale and offset parameters, while the camera poses remain fixed.
To reduce noise, we retain only the point clouds with per-pixel confidence values greater than 1.5, and we randomly downsample the point cloud by a factor of 4, as per-pixel point clouds are overly dense, which is inefficient for optimization.  


\subsection{Hyperparameters}
\label{suppl_sec:hyperparameters}
In this section, we discuss how key hyperparameters are set and evaluate their influence on performance.
We analyze the effect of the $\tau_{cos}$ in~\cref{suppl_eq:coarse_mask} and~\cref{suppl_eq:fine_mask}, and the connectivity threshold $\tau_{match}$ in~\cref{suppl_sec:object_tracking}.
Additionally, we evaluate the geometric verification threshold in Sec. 3.3 of the main paper, where we compare the chamfer distance between the two Gaussian object templates to $\tau_{overlap}$ to examine the geometric consistency across the temporal track.
We demonstrate the results of variations on above hyperparameters in~\cref{suppl_table:hyperparameters}.

For the change detection, we set $\tau_{cos}=0.9$ in~\cref{suppl_eq:coarse_mask} and~\cref{suppl_eq:fine_mask}, which were used throughout our experiments.
We sweep the values between [0.8, 0.95], referring to the cosine similarity threshold in~\cite{gescf}, originally set to 0.88.
We set $\tau_{match}=50$, considering that higher threshold values tend to neglect small objects while tracking.
In geometric verification stage, $\tau_{overlap}=0.15$ worked well to distinguish the geometrically matched objects.
Severely lower $\tau_{overlap}$ interrupts the object from being rigidly tracked.

\begin{table}[b!]
\centering
\resizebox{0.9\linewidth}{!}{
    \begin{tabular}{l|c|l|c|l|c}
        \toprule
        $\tau_{cos}$ & PSNR & $\tau_{match}$ & PSNR & $\tau_{overlap}$ & PSNR \\
        \midrule
        0.80 & 22.79 & 10 & 23.40 & 0.05 & 23.29 \\
        0.85 & 22.76 & 50 & 23.43  & 0.1 & 23.39 \\
        0.90 &  23.43 & 100 & 23.33 & 0.15 & 23.43  \\
        0.95 & 23.33 & 200 & 23.22 & 0.20 & 23.41 \\
        \bottomrule
    \end{tabular}
}
\caption{\textbf{Effects of the hyperparameter.}
}
\label{suppl_table:hyperparameters}
\end{table}

\begin{figure*}[t]
    \centering
    \includegraphics[width=\linewidth]{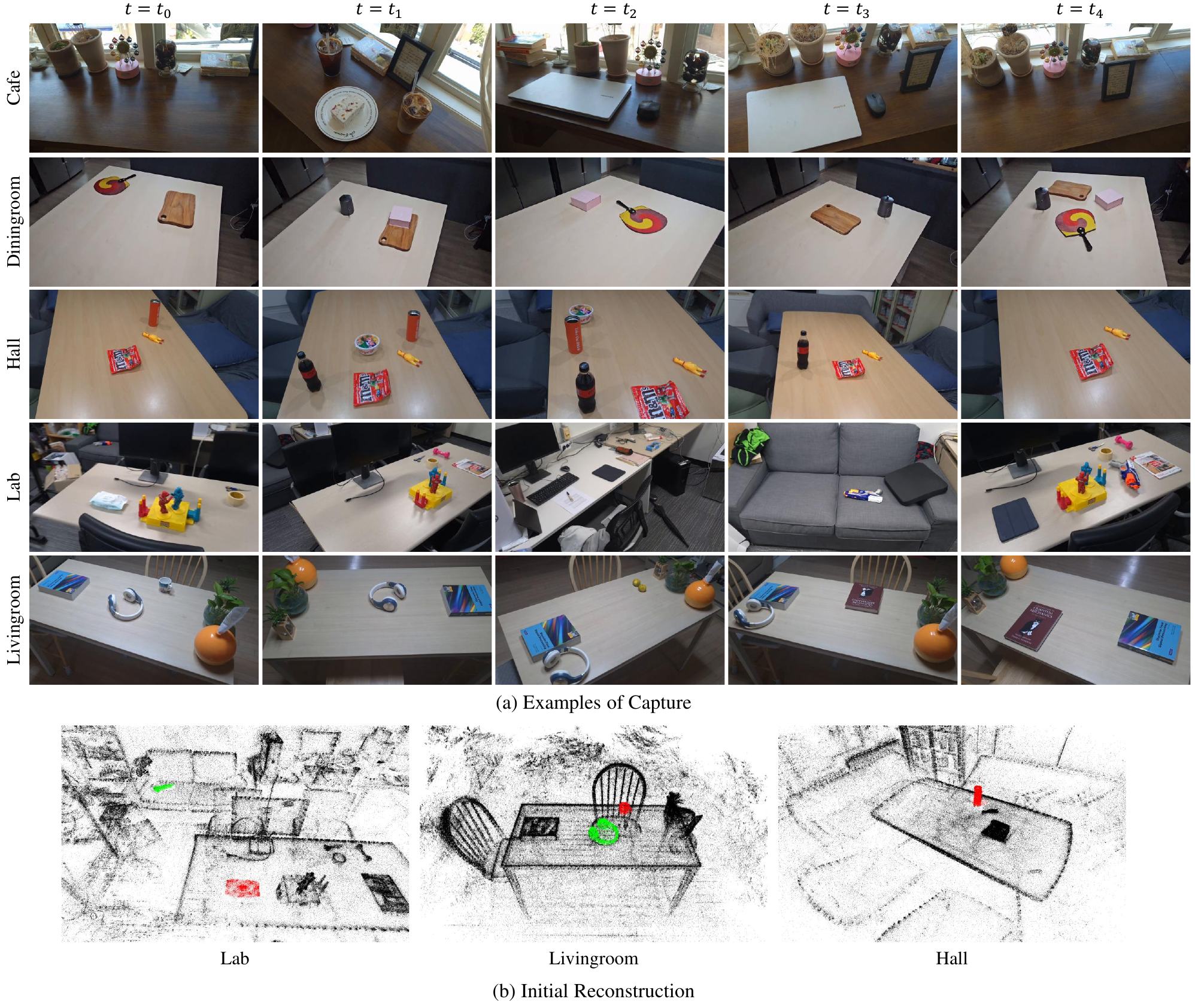}
    \caption{\textbf{Examples of our datasets.} (a) Illustration of long-term captures of our dataset for each scene. (b) Initial reconstruction of Gaussian splats at $t=0$ and tracked instances at the initial timestep.}
    \label{suppl_fig:dataset}
\end{figure*}

To initialize Gaussian primitives, we use the dense point cloud's position and its corresponding color as the initial position and color.
We convert the RGB color into spherical harmonics (SH) coefficients and initialize higher-order SH components with zeros. 
We use low initial opacity values as $\alpha=0.1$ for all Gaussians. 
Rotations are initialized as identity quaternions, i.e., $q = (1,0,0,0)$ for all Gaussians, while the scales are determined by the pairwise squared distance as done in the original 3DGS~\cite{3dgs}.
For refinement, we update the Gaussian Splats for 5000 iterations using the same learning rate as done in the official 3DGS~\cite{3dgs} implementation.  
We skipped several techniques that were used in the original 3DGS such as opacity resetting, cloning and pruning operations to preserve the original reconstruction. 

\section{Datasets}
In this section, we provide some additional details for the datasets that we have introduced in our main manuscript.
We casually captured video sequences using Galaxy S24 with fixed focal lengths and manually changed several objects between the captures for 5 sequences. 
After converting video frames to images, we regularly sampled 300-400 images for initial reconstruction to cover the scene of interest, and obtain camera poses using COLMAP SfM~\cite{colmap_sfm}.
~\cref{suppl_fig:dataset} illustrates the example of our datasets for every scene, captured from different timesteps.
In total, our dataset contains five scenes that cover a diverse set of indoor spaces (Cafe, Diningroom, Livingroom, Hall, Lab).
We additionally visualized the initial reconstruction of sampled scenes, with tracked instances using our pipeline, which serve as the initial object Gaussian templates before refinement.
For evaluation, we selected every 8th frame following the conventional evaluation protocol of neural rendering~\cite{llff, nerf}.

\section{Additional performance analysis}
\subsection{Lighting changes}
\begin{figure}[b]
    \centering
    \includegraphics[width=\linewidth]{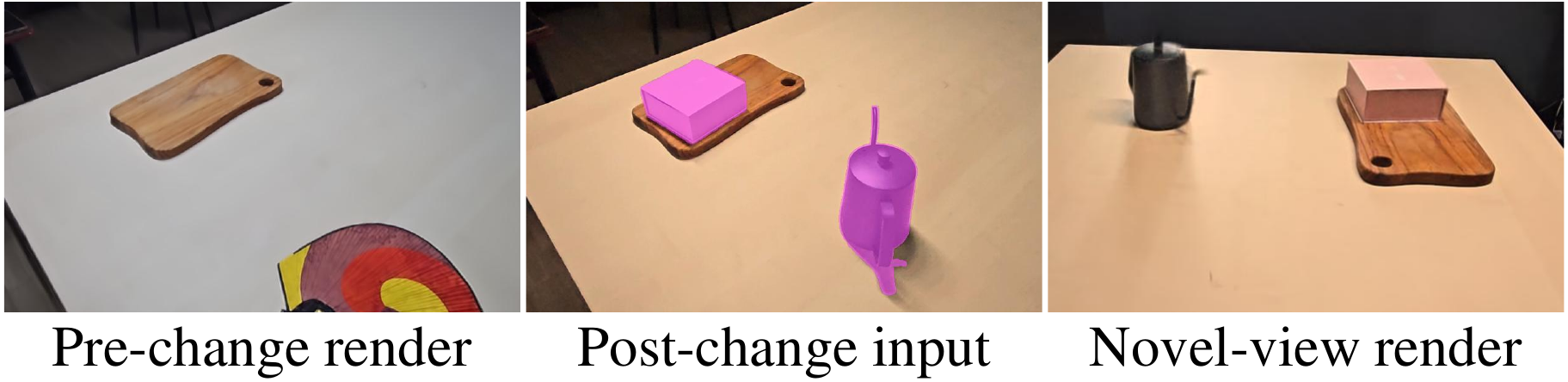}
    \caption{\textbf{Different lighting results with estimated change.}}
    \label{rebuttal_fig:lighting}
\end{figure}

To ensure robust change estimation, we use semantic differences and structural differences rather than directly comparing the RGB values. 
We validate our framework's robustness against varying illumination by applying different exposure, tone, and contrast curves to our dataset, as demonstrated in~\cref{rebuttal_fig:lighting}. 
To account for the exposure changes in the different input images, we used the exposure compensation provided in the official 3DGS~\cite{3dgs} implementation.
As a result, our framework successfully isolates object-level modifications without being degraded by global illumination shifts, maintaining high rendering fidelity.

\subsection{Geometry estimation errors}
\begin{figure}[b]
    \centering
    \includegraphics[width=\linewidth]{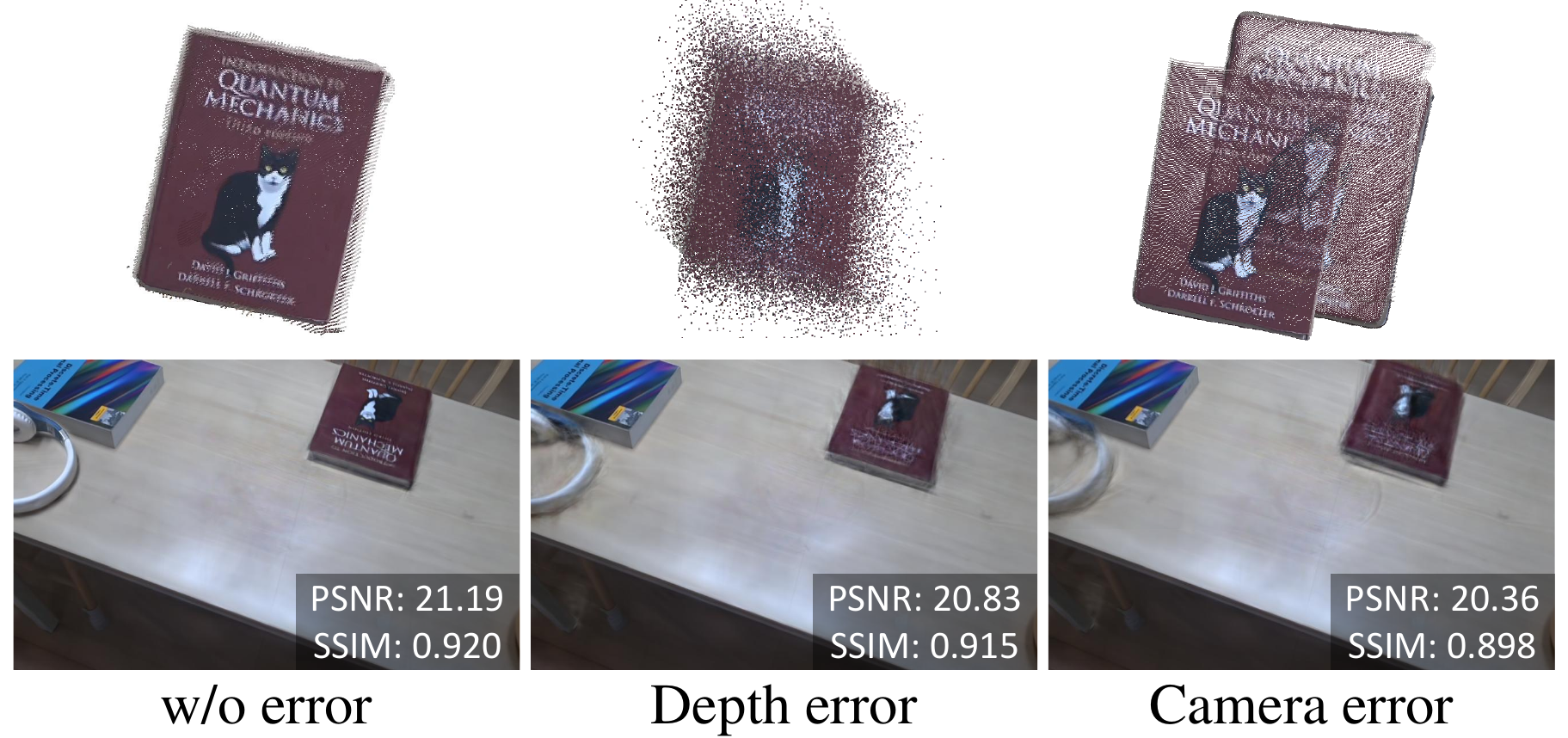}
    \caption{\textbf{Geometry estimation errors and rendering results.}}
    \label{rebuttal_fig:mast3r_error}
\end{figure}

The robustness experiments in the main paper primarily analyze descriptor noise, related to the matching error.  
To further simulate the actual estimation failures, we conduct an additional analysis by assuming errors in depth maps and camera poses during reconstruction, which closely mimics real-world failure scenarios, as depicted in~\cref{rebuttal_fig:mast3r_error}.  
While moderate noise is refined during the final refinement step, MASt3R~\cite{mast3r} point clouds with severe errors and unstable registration are naturally treated as separate instances and refined independently.
Consequently, this isolation strategy prevents initial pose or depth inaccuracies from corrupting the global scene geometry, effectively absorbing the errors to yield coherent structural reconstructions.

\subsection{Lightweight representation capability}
In our main experiments, we utilize 3 images per timestep across 4–5 timesteps per scene. 
To further evaluate the impact of input sparsity, we conduct an additional performance analysis by varying the number of images per timestep on the \textsc{Livingroom} scene, as detailed in~\cref{rebuttal_table:num_image_vs_psnr}. 
As a reference baseline, we measured the performance of standard 3DGS~\cite{3dgs} using 100 densely sampled images per timestep. Notably, our approach achieves comparable reconstruction quality using only sparse updates, eliminating the need for dense image captures.

Although we incorporate pretrained modules, our pipeline remains lightweight by leveraging reusable priors from the existing scene. 
As demonstrated in~\cref{rebuttal_table:memory_fps}, our framework effectively models dynamic scenes while maintaining real-time rendering performance after optimization.
Furthermore, it significantly reduces the memory footprint required for long-term environment maintenance, keeping memory usage comparable to the initial static 3DGS~\cite{3dgs}. 
Ultimately, these results demonstrate that our updating mechanism not only renders photorealistic images but also provides a highly scalable and memory-efficient solution for long-term scene representations.

\begin{table}[b]
    \centering
    \resizebox{\linewidth}{!}{
    \begin{tabular}{c|c|c|c|c|c|c}
         \toprule
         \# image & 2 & 3 & 4 & 5 & 6 & 3DGS~\cite{3dgs} ($\sim$100) \\
         \midrule
         PSNR & 23.06 & 23.46 & 23.92 & 24.21 & 24.41 & 25.66 \\
         Time & 6m 2s & 7m 11s & 8m 42s & 10m 4s & 11m 20s & 31m 32s \\
         Peak VRAM & 7.8 GB & 8.7 GB & 9.2 GB & 10.3 GB & 11.7 GB & 8.3 GB \\
         \bottomrule
    \end{tabular}
    }
    \caption{\textbf{Performance analysis for different number of images.}}
    \label{rebuttal_table:num_image_vs_psnr}
\end{table}

\begin{table}[b]
    \centering
    \resizebox{\linewidth}{!}{
    \begin{tabular}{c|c|c|c|c}
        \toprule
          & 3DGS~\cite{3dgs} (Single time) & 4DGS~\cite{4dgs} & CL-Splats~\cite{cl-splats} & LTGS (Ours) \\
        \midrule
         Memory & 125.8 MB & 172.8 MB & 192.9 MB & 128.5 MB \\
         FPS & 190.3 & 82.6 & 34.1 & 102.8 \\
         \bottomrule
    \end{tabular}
    }
    \caption{\textbf{Memory and dynamic rendering FPS comparison.}}
    \label{rebuttal_table:memory_fps}
\end{table}

\section{Additional comparative studies}
\subsection{Additional qualitative comparisons}
\begin{figure*}[t]
    \centering
    \includegraphics[width=0.82\linewidth]{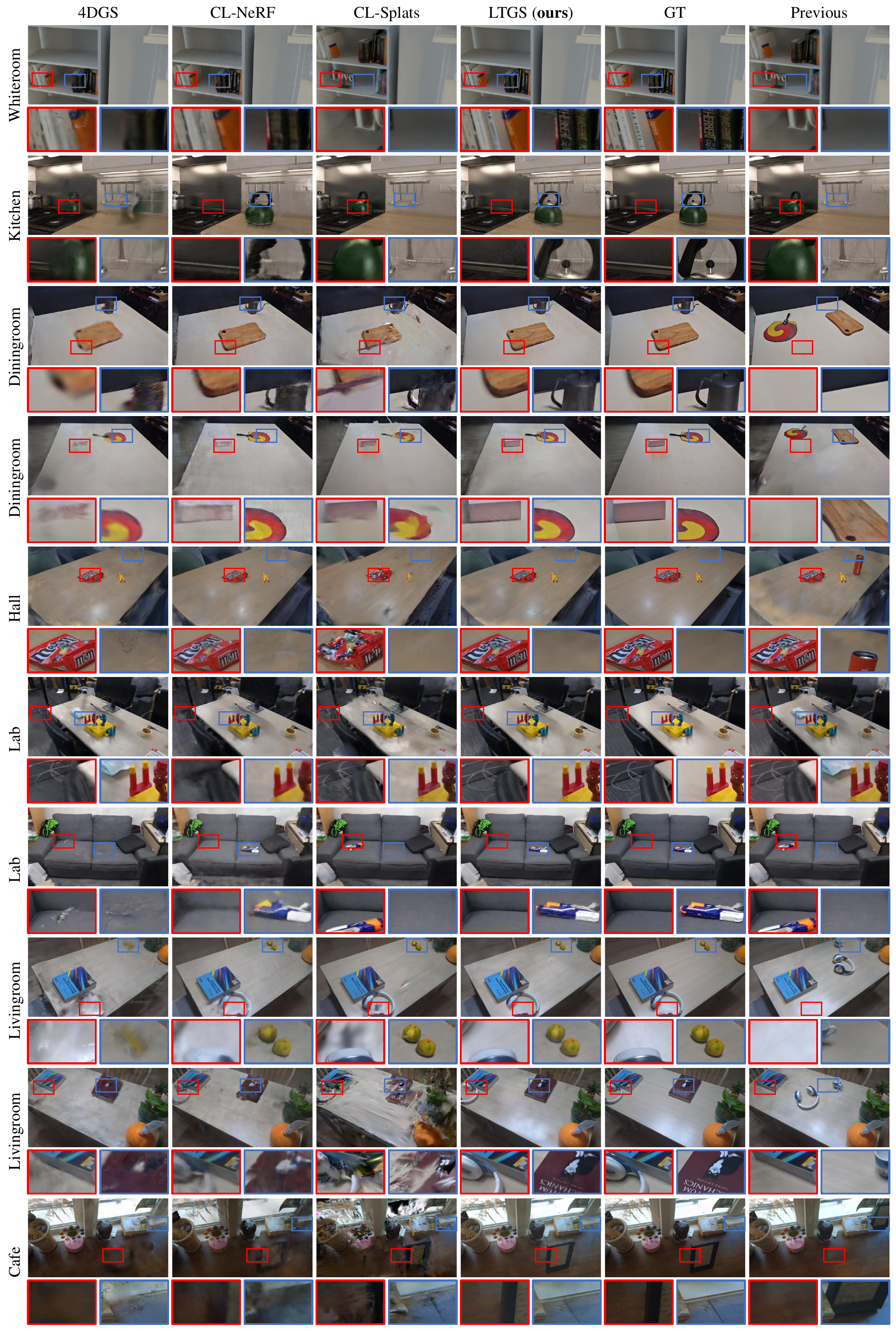}
    \caption{\textbf{Additional qualitative comparisons.} We illustrate the results of our method and baselines using CL-NeRF dataset and our dataset.}
    \label{suppl_fig:additional_qualitative}
\end{figure*}

We present additional qualitative results for both CL-NeRF~\cite{cl-nerf} and our datasets in~\cref{suppl_fig:additional_qualitative}. 
We illustrate scenes that are not covered in Fig. 3 of the main manuscript.
By incorporating reusable priors into scene representations, we achieve a notable reduction of artifacts, particularly in under-constrained regions where other baselines often struggle.
Compared to the baselines, our method produces cleaner geometry and more photorealistic synthesis under novel viewpoints.
This improvement highlights the effectiveness of our method in reducing ambiguity and enabling stable reconstructions across diverse scenes.

\subsection{Per-scene quantitative comparisons}
We provide the evaluation results of all scenes in terms of PSNR, SSIM, and LPIPS.
As demonstrated in ~\cref{suppl_table:pnsr_comparison},~\cref{suppl_table:ssim_comparison} and~\cref{suppl_table:lpips_comparison}, our framework achieves the best results for most scenes.

\begin{table*}[hb]
\small
\centering
\resizebox{0.98\linewidth}{!}{
    \begin{tabular}{l|ccc|ccccc}
	\toprule
	{} & \multicolumn{3}{c|}{CL-NeRF dataset} & \multicolumn{5}{c}{Our dataset}\\ 
        {Method} & Whiteroom & Kitchen & Rome & Cafe & Diningroom & Livingroom & Hall & Lab \\
        \midrule
        3DGS~\citep{3dgs} & 20.66 & 24.55 & \underline{28.38} & 15.97 & 18.44 & 20.44 & 22.35 & 20.60\\
        InstantSplat~\citep{instantsplat} & 19.35 & 17.66 & 19.92 & 17.71 & 22.14 & 19.30 & 18.66 & 18.98 \\
        4DGS~\citep{4dgs} & 24.84 & 25.41 & 28.15 & 18.01 & \underline{22.84} & 22.51 & \underline{23.33} & 20.76 \\
        NSC~\citep{neural_scene_chronology} & 18.13 & 17.40 & 26.36 & 15.10 & 18.21 & 19.81 & 18.63 & 15.86 \\
        3DGS-CD~\citep{3dgscd} & 23.86 & 22.20 & 24.77 & 17.94 & 22.15 & 21.00 & 22.22 & 21.41 \\
        CL-NeRF~\citep{cl-nerf} & 26.22 & \underline{25.81} & 24.55 & 16.53 & 22.66 & 22.32 & 22.40 & 20.87 \\
        CL-Splats~\citep{cl-splats} & \textbf{26.84} & 24.90 & 25.79 & \underline{18.71} & 19.29 & \textbf{23.94} & 22.21 & \underline{21.47} \\
        \midrule
        LTGS (ours) & \underline{26.67} & \textbf{26.21} & \textbf{28.65} & \textbf{20.77} & \textbf{25.22} & \underline{23.50} & \textbf{25.22} & \textbf{22.64} \\
        \bottomrule
    \end{tabular}
}
\caption{\textbf{PSNR comparisons on CL-NeRF dataset and our dataset.} The first and second best results are highlighted in \textbf{bold} and \underline{underlined}, respectively.}
\label{suppl_table:pnsr_comparison}
\end{table*}
\begin{table*}[hb]
\small
\centering
\resizebox{0.98\linewidth}{!}{
    \begin{tabular}{l|ccc|ccccc}
	\toprule
	{} & \multicolumn{3}{c|}{CL-NeRF dataset} & \multicolumn{5}{c}{Our dataset}\\ 
        {Method} & Whiteroom & Kitchen & Rome & Cafe & Diningroom & Livingroom & Hall & Lab \\
        \midrule
        3DGS~\citep{3dgs} & 0.821 & \textbf{0.645} & \textbf{0.899} & \underline{0.776} & 0.866 & \underline{0.897} & \underline{0.911} & \underline{0.835} \\
        InstantSplat~\citep{instantsplat} & 0.699 & 0.443 & 0.660 & 0.723 & 0.857 & 0.795 & 0.809 & 0.739 \\
        4DGS~\citep{4dgs}  & 0.827 & 0.644 & 0.885 & 0.772 & \underline{0.893} & 0.882 & 0.892 & 0.812 \\
        NSC~\citep{neural_scene_chronology} & 0.710 & 0.536 & 0.849 & 0.705 & 0.795 & 0.813 & 0.804 & 0.658 \\
        3DGS-CD~\citep{3dgscd} & 0.815 & 0.563 & 0.803 & 0.719 & 0.792 & 0.797 & 0.751 & 0.812 \\
        CL-NeRF~\citep{cl-nerf} & 0.829 & 0.636 & 0.725 & 0.696 & 0.863 & 0.881 & 0.859 & 0.775 \\
        CL-Splats~\citep{cl-splats} & \textbf{0.848} & 0.627 & 0.840 & 0.749 & 0.873 & 0.836 & 0.866 & 0.819\\
        \midrule
        LTGS (ours) & \textbf{0.848} & \textbf{0.645} & \underline{0.892} & \textbf{0.845} & \textbf{0.911} & \textbf{0.922} & \textbf{0.924} & \textbf{0.840} \\
        \bottomrule
    \end{tabular}
}
\caption{\textbf{SSIM comparisons on CL-NeRF dataset and our dataset.} The first and second best results are highlighted in \textbf{bold} and \underline{underlined}, respectively.}
\label{suppl_table:ssim_comparison}
\end{table*}
\begin{table*}[hb]
\small
\centering
\resizebox{0.98\linewidth}{!}{
    \begin{tabular}{l|ccc|ccccc}
	\toprule
	{} & \multicolumn{3}{c|}{CL-NeRF dataset} & \multicolumn{5}{c}{Our dataset}\\ 
        {Method} & Whiteroom & Kitchen & Rome & Cafe & Diningroom & Livingroom & Hall & Lab \\
        \midrule
        3DGS~\citep{3dgs} & 0.522 & 0.537 & \textbf{0.116} & \underline{0.323} & 0.313 & \underline{0.216} & \underline{0.234} & \underline{0.273} \\
        InstantSplat~\citep{instantsplat} & 0.561 & 0.579 & 0.257 & 0.328 & \underline{0.295} & 0.367 & 0.377 & 0.346 \\
        4DGS~\citep{4dgs} & 0.539 & 0.547 & 0.148 & 0.363 & 0.306 & 0.307 & 0.299 & 0.334 \\
        NSC~\citep{neural_scene_chronology} & 0.585 & 0.619 & 0.192 & 0.441 & 0.385 & 0.426 & 0.431 & 0.510 \\
        3DGS-CD~\citep{3dgscd} & 0.527 & 0.571 & 0.213 & 0.337 & 0.366 & 0.345 & 0.375 & 0.318 \\
        CL-NeRF~\citep{cl-nerf} & 0.521 & \underline{0.536} & 0.339 & 0.463 & 0.324 & 0.328 & 0.350 & 0.428 \\
        CL-Splats~\citep{cl-splats} & \underline{0.492} & 0.542 & 0.214 & 0.345 & 0.308 & 0.323 & 0.303 & 0.280 \\
        \midrule
        LTGS (ours) & \textbf{0.478} & \textbf{0.522} & \underline{0.128} & \textbf{0.246} & \textbf{0.253} & \textbf{0.185} & \textbf{0.204} & \textbf{0.260} \\
        \bottomrule
    \end{tabular}
}
\caption{\textbf{LPIPS comparisons on CL-NeRF dataset and our dataset.} The first and second best results are highlighted in \textbf{bold} and \underline{underlined}, respectively.}
\label{suppl_table:lpips_comparison}

\end{table*}

\clearpage
\end{document}